\documentclass[twocolumn,3p,times,procedia]{elsarticle}

\usepackage{ecrc}
\usepackage{graphicx}
\usepackage{balance}
\usepackage{array,threeparttable}
\usepackage{url}

\usepackage{graphicx}%
\usepackage{multirow}%
\usepackage{amsmath,amssymb,amsfonts}%
\usepackage{amsthm}%
\usepackage{mathrsfs}%
\usepackage[title]{appendix}%
\usepackage{xcolor}%
\usepackage{textcomp}%
\usepackage{manyfoot}%
\usepackage{booktabs}%
\usepackage{algorithm}%
\usepackage{algorithmicx}%
\usepackage{algpseudocode}%
\usepackage{listings}%
\usepackage{float}
\usepackage{subfig}
\usepackage{hyperref}

\graphicspath{{fig/}}
\newtheorem{myDef}{Definition}

\volume{00}

\firstpage{1}

\runauth{}

\jid{procs}

\CopyrightLine{2022}{Published by Elsevier Ltd.}

\usepackage{amssymb}
\usepackage{amsmath}
\usepackage{multicol}
\usepackage{algorithm}
\usepackage{algpseudocode}

\usepackage[figuresright]{rotating}
\usepackage{bm}
\usepackage{caption}
\usepackage{titlesec}
\titleformat*{\section}{\small \bf}
\titleformat*{\subsection}{\small \em}
\titleformat*{\subsubsection}{\small \em}

\usepackage{fancyhdr}
\fancyheadoffset[RO,EL]{0pt}
\usepackage{geometry}
\begin{document}\small
\begin{frontmatter}




\dochead{}
\title{
\begin{flushleft}
{\LARGE CGGM: A conditional graph generation model with adaptive sparsity for node anomaly detection in IoT networks} 
\end{flushleft}
}
 %

\author[]{\leftline {Munan Li$^{a,1}$, Xianshi Su$^{a,1}$, Runze Ma$^a$, Tongbang Jiang$^a$, Zijian Li$^{a,*}$, Tony Q.S. Quek$^{b,c}$ \textit{Fellow,IEEE}}}

\address{ \leftline {$^a$College of Artificial Intelligence, Dalian Maritime University, China}
  
  \leftline {$^b$Information Systems Technology and Design, Singapore University of Technology and Design, Singapore}

  \leftline {$^c$Yonsei Frontier Lab, Yonsei University, South Korea }
}

\cortext[]{Corresponding author. College of Artificial Intelligence, Dalian Maritime University, China.\\
E-mail address: limunan@dlmu.edu.cn(M.Li), suxianshi15@gmail.com(X,Su),\\
mrz@dlmu.edu.cn(R.Ma), jtb@dlmu.edu.cn(T.Jiang),\\
lizj@dlmu.edu.cn(Z.Li), tonyquek@sutd.edu.sg(T.Quek)}
\fntext[equal]{The two authors contribute equally to this work.}

\begin{abstract}

Dynamic graphs are extensively employed for detecting anomalous behavior in nodes within the Internet of Things (IoT). Graph generative models are often used to address the issue of imbalanced node categories in dynamic graphs. Nevertheless, the constraints it faces include the monotonicity of adjacency relationships, the difficulty in constructing multi-dimensional features for nodes, and the lack of a method for end-to-end generation of multiple categories of nodes. In this paper, we propose a novel graph generation model, called CGGM, specifically for generating samples belonging to the minority class. The framework consists two core module: a conditional graph generation module and a graph-based anomaly detection module. The generative module adapts to the sparsity of the matrix by downsampling a noise adjacency matrix, and incorporates a multi-dimensional feature encoder based on multi-head self-attention to capture latent dependencies among features. Additionally, a latent space constraint is combined with the distribution distance to approximate the latent distribution of real data. The graph-based anomaly detection module utilizes the generated balanced dataset to predict the node behaviors. Extensive experiments have shown that CGGM outperforms the state-of-the-art methods in terms of accuracy and divergence. The results also demonstrate CGGM can generated diverse data categories, that enhancing the performance of multi-category classification task.

\end{abstract}


\end{frontmatter}


\section{Introduction}
\label{intro}
Anomaly detection is an important tool for intrusion detection systems (IDS)~\cite{kale2023few}. An efficient network security mechanism not only need to identify abnormal traffic but also to analyze the behavior of the nodes generating such traffic. One illustrative method is to construct a dynamic graph called Traffic Dispersion Graphs (TDG)~\cite{iliofotou2009exploiting} to represent the distribution of network traffic within specific time intervals and utilize graph-based models to perform anomaly detection. However, the existed graph-based anomaly detection algorithms predominantly belong to supervised learning, requiring labeled types for nodes~\cite{lo2023xg}. In practical scenarios, there is a noticeable imbalance observed in traffic samples~\cite{douzas2018effective}, resulting in a reduced sensitivity of the detection model towards minority classes~\cite{adiban2023step}.
\par
Generative models can effectively tackle this issue~\cite{simonovsky2018graphvae,bojchevski2018netgan}. There are still several key challenges to be addressed to implement a graph generation model in anomaly detection. First, existing graph generation methods often lack flexibility. When generating the inherent distribution of nodes, existing methods such as TableGAN~\cite{park2018data} and CTGAN~\cite{xu2019modeling} can only utilize the relationships between nodes while ignoring the attribute features. This will lead to a simplistic topology for the adjacency matrix, lacking adaptability to the complexity and variability of traffic data. Secondly, IoT traffic data often exhibits multi-dimensional attribute features\cite{atwood2017sparse}, such as node transceiver link relationships and category labels. Existing studies primarily focus on link prediction task that model single-dimensional feature~\cite{lei2019gcn,guo2019deep}. These approaches fail to generate rich data patterns~\cite{simonovsky2018graphvae} and obtain the latent dependency between multiple features~\cite{you2018graphrnn}. Finally, it is difficult to distinguish real data from fake data from a higher-dimensional semantic perspective. Current loss calculation methods typically consider the loss between individual data points from a lower-dimensional space, and ignore the distribution distance between structural features.

To this end, we introduce a GAN-based conditional graph generation model (CGGM) for generating traffic graph snapshots, aiming to achieve a better data balance. CGGM integrates the topological structure of traffic networks with the multi-dimensional features of nodes to model the evolution of graph snapshots. Specifically, CGGM first propose an adaptive sparsity adjacency matrix generator to refine the construction of adjacency matrix. It leverages the sparsity observed in real adjacency matrices by down-sampling a noise adjacency matrix to match the sparsity level of real data. Additionally, it incorporates the node attributes from traffic dispersion graphs to capture the inherent distribution of network traffic. Then, to ensure the generated features encompass rich data patterns, we incorporate a self-attention based multi-dimensional feature encoder into the generator and discriminator training processes and capture the latent dependencies among these features. Furthermore, label embedding integrated as a conditional constraint to achieve the controlled generation of nodes with specific category and corresponding features. Finally, we propose a latent space constraint in the generator by calculating the distance between generated embeddings in the high-dimensional latent space. This constraint will be combined with the distribution distance between true and generated samples to approximate the latent distribution of real data. Extensive experiments demonstrate that CGGM outperforms the state-of-the-art baselines in terms of both data distribution similarity, and classification performance.
  
The main contributions of this paper are summarized as follows:
\begin{enumerate}[(1)]

\item[$\bullet$] We propose an adaptive sparsity adjacency matrix generator that refines adjacency matrix construction to match the sparsity of real data and incorporate node attributes from traffic dispersion graphs to accurately capture the network traffic distribution.
\item[$\bullet$] We develop a multi-dimensional feature encoder with self-attention mechanism to generate features with rich data pattern and capture their dependencies. Additionally, we enhance the encoder by preserving topology, attributes, and label context to control the generation of nodes with semantic relevance to the labels.

\item[$\bullet$] A latent space constraint is incorporated into the generator and combined with distribution distance from real data to better constrain sample generation, which approximate the latent distribution of real data.

\end{enumerate}

The rest of the paper is organised as follows. \autoref{sec:relatedwork} describes the work related to graph generation model and GNN-based anomaly detection. \autoref{sec:preliminary} provides the definition of data structure and tasks. In \autoref{sec:admodel}, the methodology for constructing dynamic traffic graphs is presented, and the anomaly node detection method and its key modules is described. \autoref{sec:experiment} describe the experimental setup and analysis of the experimental results. Finally, in \autoref{sec:conclusion}, conclusions and prospects for future work are provided.

\section{Related Works}
\label{sec:relatedwork}

In this section, we briefly summarise IoT anomaly detection and related work, and analyse the generative model for graph structures.

\subsection{Anomaly Detection based on Graph Neural Networks}
Network traffic samples are usually represented as time series, where each row represents the communication between network nodes. The first step in analysing the information within the graph is to transform the traffic data into a graph structure~\cite{zola2022network}.~\cite{iliofotou2009exploiting} proposed a Traffic Dispersion Graph (TDG) concept which splits time-series traffic into subintervals with fixed time intervals and extracts a graph snapshot from each subinterval.

Graph Neural Networks (GNN) have emerged as a promising method for anomaly detection~\cite{wang2021decoupling}. \cite{wang2022contrastive} propose an Contrastive GNN-based traffic anomaly analysis for imbalanced datasets in an IoT-based intelligent transport system.~\cite{duan2022application} proposed an intrusion detection method based on semi-supervised learning of Dynamic Line Graph Neural Networks (DLGNN).~\cite{wu2021graph} provide an in-depth study of graph neural networks (GNNs) for anomaly detection in smart transport, smart energy and smart factories.

\subsection{Graph Generation Model}
Generate network traffic through GAN has become a common method in the field of anomaly detection ~\cite{ring2019flow}. Recently, deep generative models of graphs have been applied to anomaly detection, biology, and social sciences~\cite{guo2022deep}. There are many techniques to construct virtual features of graph node data from different perspectives~\cite{wang2018graphgan,nauata2020house,chang2021building}.
For example,~\cite{guo2023regraphgan} proposed a new graph generation adversarial network to solve the problem of encoding complex in dynamic graph data.~\cite{guo2022deep} proposed a graph-translation-generative-adversarial-nets (GT-GAN) model. Models such as TableGAN~\cite{park2018data} and CTGAN~\cite{xu2019modeling} are utilized for constructing tabular data. In addition, some methods are trained to link prediction~\cite{lei2019gcn,guo2019deep,simonovsky2018graphvae}. 

\section{Problem Formulation}\label{sec:preliminary}
In this section, we present the relevant graph structure for graph generation tasks. In the meantime, we provide a formal definition of a graph generation task.

\begin{myDef}\label{def:real} Real Graph. We define the real graph be $G_r =({\boldmath{V_r}},{\boldmath{A_r}},{\boldmath{X_r}},{\boldmath{C_r}})$, where ${\boldmath{V_r}}$ is the set of nodes, ${\boldmath{A_r}}\in{\mathbb{R}^{N \times N}}$ is the adjacency matrix, and ${\boldmath{X_r}}\in{\mathbb{R}^{N\times F}}$ is the feature matrix, where each row denotes a node feature vector $x_i$, $a_{ij}\in{{\boldmath{A}}}$ denotes the weights of the edges between the nodes $v_i$ and $v_j$, and ${\boldmath{C_r}}\in{\mathbb{R}^{N}}$ denotes the category label of nodes.
\end{myDef}

\begin{myDef}\label{def:original} Noisy Graph. Let the noisy graph be $G_o =({\boldmath{V_o}},{\boldmath{A_o}},{\boldmath{X_o}},{\boldmath{C_o}})$, where ${\boldmath{X}}_o\in{\mathbb{R}^{N \times F}}$, ${C_o}\in{\mathbb{R}^{N}}$ and $A_o \in {\mathbb{R}^{N\times N}}$. It is generated form random noise.
\end{myDef}

\begin{myDef}\label{def:target} Synthetic Graph. Similarly, we define the synthetic graph as $G_g=({\boldmath{V_t}},{\boldmath{A_g}},{\boldmath{X_g}},{\boldmath{C_g}})$, which shares a similar data distribution to $G_r$.
\end{myDef}
\textbf{Problem Formulation.} Based on the aforementioned definitions, the task of graph generation is formulated as follows: \textbf{Input:} the noisy graph $G_{o}$, and the real graph $G_r$. \textbf{Output:} A graph generation model with a mapping function $\mathcal{F}(\cdot)$ that transform the noisy graph $G_{o}$ into a synthetic graph $G_{g}$ with specific data distribution similar to $G_r$. By generating synthetic graphs $G_{g}$, we can create a balanced dataset for downstream anomaly detection task.\section{Problem Formulation}\label{sec:preliminary}
In this section, we present the relevant graph structure for graph generation tasks. In the meantime, we provide a formal definition of a graph generation task.

\begin{myDef}\label{def:real} Real Graph. We define the real graph be $G_r =({\boldmath{V_r}},{\boldmath{A_r}},{\boldmath{X_r}},{\boldmath{C_r}})$, where ${\boldmath{V_r}}$ is the set of nodes, ${\boldmath{A_r}}\in{\mathbb{R}^{N \times N}}$ is the adjacency matrix, and ${\boldmath{X_r}}\in{\mathbb{R}^{N\times F}}$ is the feature matrix, where each row denotes a node feature vector $x_i$, $a_{ij}\in{{\boldmath{A}}}$ denotes the weights of the edges between the nodes $v_i$ and $v_j$, and ${\boldmath{C_r}}\in{\mathbb{R}^{N}}$ denotes the category label of nodes.
\end{myDef}

\begin{myDef}\label{def:original} Noisy Graph. Let the noisy graph be $G_o =({\boldmath{V_o}},{\boldmath{A_o}},{\boldmath{X_o}},{\boldmath{C_o}})$, where ${\boldmath{X}}_o\in{\mathbb{R}^{N \times F}}$, ${C_o}\in{\mathbb{R}^{N}}$ and $A_o \in {\mathbb{R}^{N\times N}}$. It is generated form random noise.
\end{myDef}

\begin{myDef}\label{def:target} Synthetic Graph. Similarly, we define the synthetic graph as $G_g=({\boldmath{V_t}},{\boldmath{A_g}},{\boldmath{X_g}},{\boldmath{C_g}})$, which shares a similar data distribution to $G_r$.
\end{myDef}

\textbf{Problem Formulation.} Based on the aforementioned definitions, the task of graph generation is formulated as follows: \textbf{Input:} the noisy graph $G_{o}$, and the real graph $G_r$. \textbf{Output:} A graph generation model with a mapping function $\mathcal{F}(\cdot)$ that transform the noisy graph $G_{o}$ into a synthetic graph $G_{g}$ with specific data distribution similar to $G_r$. By generating synthetic graphs $G_{g}$, we can create a balanced dataset for downstream anomaly detection task.

\section{Method}\label{sec:admodel}
\begin{figure*}[h]
  \centering
  \includegraphics[width=\linewidth]{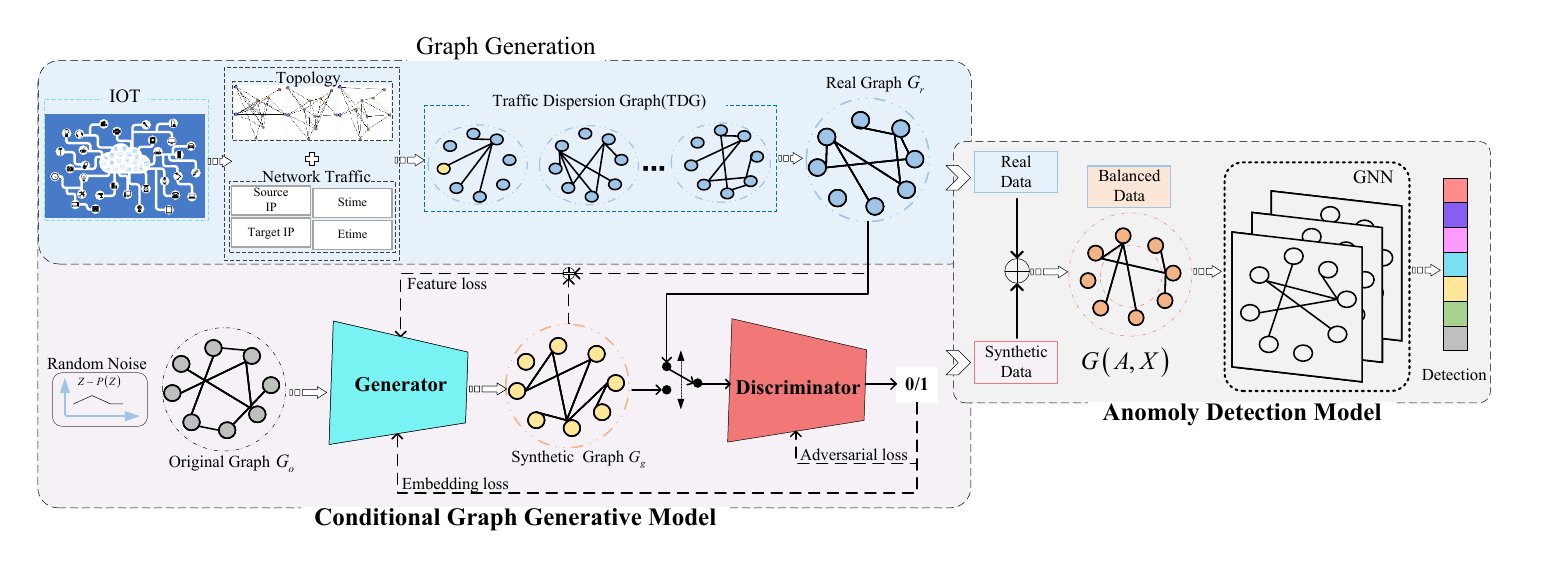}
  \caption{The overall framework of the process. The framework is made up of three components: Graph Generation, Conditional Graph Generative Model and Anomaly Detection Model. Firstly, a sequence of graph snapshots is extracted from the traffic samples by Traffic Dispersion Graphs(TDG) construction method. The conditional graph generation model is then used to generate synthetic data that approximates the real data. Eventually, the synthetic data is aggregated with the real data as an input to the anomaly detection model. By capturing the spatial structure features of the nodes, the anomaly detection model can predict anomalies.}
 \label{fig:TGAT_framwork}
\end{figure*}

In this section, a detailed description of the anomaly detection framework based on graph generation model is presented. Specifically, the framework analyses the traffic evolution characteristics of the traffic dispersion graph to generate samples with a real distribution and capture complex anomaly data patterns. It mainly consists of two core components: the conditional graph generative module, which synthesizes the graph snapshot data, the anomaly detection module, which detects anomaly data based on graph neural network. The overall framework is shown in Figure \ref{fig:TGAT_framwork}.

\subsection{Traffic Dispersion Graph Generation}
The purpose of the Traffic Dispersion Graph (TDG)~\cite{iliofotou2007network} is to extract a sequence of graph snapshots containing spatio-temporal evolutionary information from network traffic. Each traffic sample is represented by a $3$-tuple of \{\textit{source node IP},\textit{ target node IP}, \textit{traffic characteristics}\}, where each IP address is represented as a node, and each traffic is regarded as a group of communication interactions between nodes. 
 
Algorithm \ref{alg:graph_data} generates a real graph that reflects the network communication patterns from network traffic data. The adjacency matrix ${\boldmath{A}}_{r}$ of the snapshot is obtained by traversing the IP address, the $a_{ij}$ is corresponded to the source IP address and the destination IP address. The feature matrix ${\boldmath{X}}_{r}$ of the graph snapshot is obtained by aggregating traffic features, and the label matrix ${\boldmath{C}}_{r}$ of the graph snapshot is obtained by counting Label in the traffic features. The feature $x\in{\mathbb{R}^F}$ of each node $v$ is generated from the weighted average of the $F$-dimensional traffic features, and the feature matrix of the obtained real graph $G_{r}$ is $ {X}_{r}=\{x_1,x_2, \ldots, x_N \} \in {\mathbb{R}^{N\times F}}$. The label vector is defined as ${\boldmath{C}}_{r}=\{c_1,c_2, \ldots, c_N \}\in{\mathbb{R}^{N}}$, where $c_i \in \{0,1\}$ represents whether the node is normal or abnormal, e.g. $c_i=1$ representing abnormal.

\floatname{algorithm}{Algorithm}
\begin{algorithm}[ht!] 
\footnotesize
	\renewcommand{\algorithmicrequire}{\textbf{Input:}}
	\renewcommand{\algorithmicensure}{\textbf{Output:}}
	\caption{Constructing the TDG}  
	\label{alg:graph_data}
	\begin{algorithmic}[1] 
		\Require Cybersecurity data set $D$, Time interval $K$, IP dictionary $S$, Source IP column $IP_s$, Destination IP column $IP_d$, Label column $Label$.
		\Ensure Real Graph snapshot $G_{r}$.
               \State $k\gets{}$0
				    \For{each $i=1:Len(D)$}
                        \If{$k<K$}
					 \If{$\text{S}[IP_s[i]]$ and $\text{S}[IP_d[i]]$}
						\State${\boldmath{A}}[IP_s[i]][IP_d[i]]\gets{1}$
                            \State ${\boldmath{X}}\gets{D[i]}$
                            \State ${\boldmath{E}}\gets{Label[i]}$
					\EndIf
      \Else \State $G\gets{[{\boldmath{A}},{\boldmath{X}},{\boldmath{E}}]}$
      \State $k\gets{}$0
      \EndIf
				\EndFor	
 \State \Return{ $G_{r}$ }
	\end{algorithmic}
\end{algorithm}

\subsection{Conditional Graph Generative Model}
\begin{figure*}[h]
  \centering
  \includegraphics[width=0.8\linewidth]{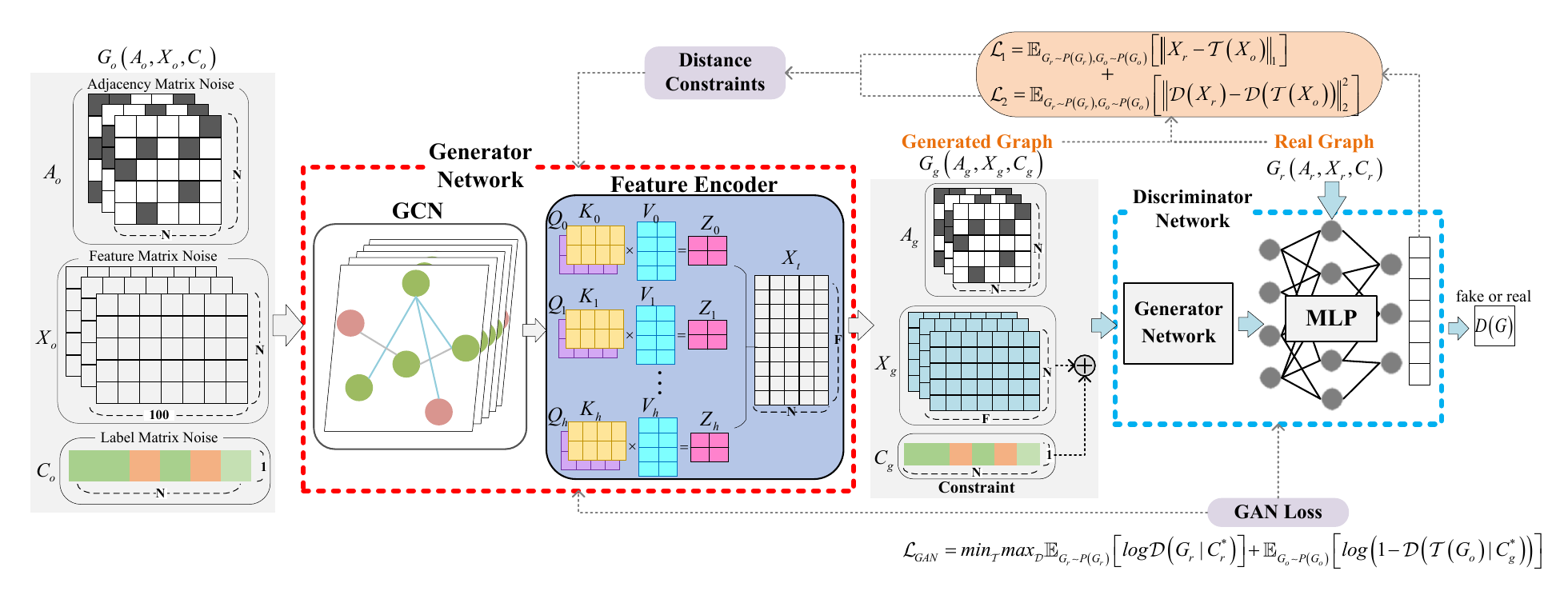}
  \caption{A detailed illustration of the CGGM. The model consists of a generator network $\mathcal{T}$ and a discriminator $\mathcal{D}$. $\boldmath{G_o}$ is the input random noise graph data, $\boldmath{G_r}$ is the real graph data, and $\boldmath{G_g}$ is the synthetic graph data generated by the model.}
 \label{fig:TGAT}
\end{figure*}

The structure of the graph generation model is shown in Figure \ref{fig:TGAT}. It consists of two main components: the condition graph generator $\mathcal{T}$ and the discriminator $\mathcal{D}$.

\subsubsection{Condition Generator Network}
The conditional graph generator consists of a convolutional neural network (GCN) and an feature generation module. We use the GCN to construct the local topology of each graph snapshot. A typical GCN unit takes the feature matrix ${\boldmath{X}}$ as input and performs a local first-order approximation of the spectrogram convolution operation, which is defined as:
\begin{equation}
   {\rm GCN}\left({\boldmath{A}},{\boldmath{X}}\right)=\sigma\left(\hat{{\boldmath{D}}}^{-1/2}\hat{{\boldmath{A}}}\hat{{\boldmath{D}}}^{-1/2}XW\right)
    \label{eq:f1}
\end{equation}
Here, $\hat{{\boldmath{D}}}^{-1/2}\hat{{\boldmath{A}}}\hat{{\boldmath{D}}}^{-1/2}$ is the approximate graph convolution filter; $\hat{{\boldmath{A}}}={\boldmath{A}}+{\boldmath{I}}_N$,${\boldmath{I}}_N$ is the $N$-dimensional unitary matrix; ${\boldmath{D}}$ is the degree matrix; ${\boldmath{W}}$ denotes the learnable weighting matrix; The activation function, denoted as $\sigma(\cdot)$, introduces non-linearity to data representation while normalizing the value within the range of $[0,1]$. Given the input feature matrix noise ${\boldmath{X}}_o\in{\mathbb{R}^{N\times F}}$, the label noise vector ${C_o}\in{\mathbb{R}^{N}}$, and the adjacency matrix noise ${\boldmath{A}}_o\in{\mathbb{R}^{N\times N}}$, the GCN unit first takes (${\boldmath{A}}_o,{\boldmath{X}}_o$) as input and update the node representations ${\boldmath{X}}_{o}^{*}$ as:
\begin{equation}
     {\boldmath{X}}_{o}^{*}= GCN\left({\boldmath{A_o}},{\boldmath{X_o}}\right)
     \label{eq:gcn}
\end{equation}

The noise values of the noise are generated according to a certain probability distribution $p$ (e.g., uniform distribution). Then, the hidden representation ${\boldmath{X}}_{o}^{*}$ are fed into an self-attention based module to capture the feature dependencies.

{\textbf{Sparsity adjustment}}. Adjacency matrix sparsity refers to the proportion of zero elements in a matrix to the total number of elements. To generate adjacency matrices with similar sparsity to real data, we propose an adaptive sparsity-based adjacency matrix generation mechanism. Specifically, we detect the sparsity of the adjacency matrix  from real data and down-sample the noise in the adjacency matrix ${\boldmath{A}}_o$ to guide the generation of sparse matrices ${\boldmath{A}}_g$.

Since our goal is to balance the proportion of categories, it is necessary to generate sufficient category labels for each class. To achieve this propose, we first use ${\boldmath{C}}_o$ to calculate the category proportion of the original graph labels. Then, we calculate the minimum quantity required to balance each category. The label matrix ${\boldmath{C}}_t$ for the synthetic graph can be generated based on these quantity.

{\textbf{The multi-dimensional feature generator}}. To ensure that the generated feature matrix effectively reflects the graph's structure and properties, we propose a multi-dimensional feature generator based on self-attention. Specifically, we utilize a Transformer-like attention mechanism to learn the dependencies among node features. The node embedding matrix ${\boldmath{X}}_{o}^{*}$ will first put into a linear transformation layer with learning parameters $\left \{W_{q},W_{k},W_{v} \right \} $ to generate the query, key, and value matrices $Q$, $K$, and $V$. i.e. $Q=W_{q}{\boldmath{X}}_{o}^{*}$. A similar operation is employed to get $K$and $V$. Then, the multi-dimensional feature correlation encoder is defined based on scaling dot-product attention, which is formulated as:

\begin{equation}
     {\rm{Attn}}\left({\boldmath{Q}},{\boldmath{K}},{\boldmath{V}}\right)={\rm{softmax}}(\frac{{\boldmath{Q^{T}K}}}{\sqrt{F}}){\boldmath{V}}.
         \label{eq:f2}
\end{equation}
where the softmax activation function is used to normalize the attention weights, allowing us to compress the matrix ${\boldmath{V}}$ into a smaller representative embedding to simplify inference in downstream neural network operations. $\sqrt{F}$ is the weight scaling factor, which reduce the variance of the weights during training and promote stable training. Motivated by the multi-head self-attention mechanism~\cite{vaswani2017attention}, we further map ${\boldmath{Q}}$, ${\boldmath{K}}$, ${\boldmath{V}}$ to different subspace. The $h$-th head projection matrix is defined as $\left\{{\boldmath{Q}}_h,{\boldmath{K}}_h,{\boldmath{V}}_h\right\} \in{\mathbb{R}^{N\times F/h}}$, where $h$ is the number of heads. The embedding matrix ${Z'}\in{\mathbb{R}^{N \times F}}$ for the generated graph can further refined by the multi-head attention as follows:
\begin{equation}
   \begin{aligned}
 {\boldmath{Z}}={\Vert_{h=1}^{H}}\left(Z_1,\ldots,Z_H\right),\\
 Z_h={\rm{Attn}}\left(\boldmath{Q}_h,{\boldmath{K}}_h,{\boldmath{V}}_h\right).
   \end{aligned} 
    \label{eq:f3}
\end{equation}

Afterwards, we defined an embedding layer $\Upsilon (\cdot)$ with learnable parameter matrix $\left \{W_{em},b_{em}\right \} $, which can be realized as a multilayer perceptron. The input label noise $C_o$ will be put into $\Upsilon (\cdot)$ to obtain the label's specific embedding. Then the conditional information constraint $\Upsilon (C_o)$ based on label information can be defined as:
\begin{equation}
    C_g^{*} = ReLU(\Upsilon (C_o;W_em,b_em)).
    \label{eq:cons}
\end{equation}
The conditional constraint $C_o^{*}$ will be accumulated to the $\boldmath{Z'}$. The conditional constraint $C_r^{*}$ and $C_g^{*}$ can be obtained in an analogous way. Finally, We can obtain the feature matrix $\boldmath{X_g}$ for the generated target graph as ${\boldmath{X_g}}={Z}+C_g^{*}$.

\subsubsection{Discriminator Network}
We use a network that mirrors the structure of the generator network $\mathcal{T}$ as the discriminator network $\mathcal{D}$, with a fully connected network as the output layer. In the discriminator $\mathcal{D}$, we first concatenate the label constraint $C^{*}\in\left\{C_r^{*},C_g^{*}\right\}$ with the feature matrix $X\in\left\{X_r,X_g\right\}$ to generate the input feature $T$, using the formula: $\boldmath{T}=concate\left(X,C^{*}\right)$. Then, it will be map into the generator network through a fully connected layer, which is presented as follows:
\begin{equation}
\label{eq:f5}
\mathcal{D}(T)=\left(\left(\mathcal{T}(T)W_1+b_1\right)W_2+b_2\right)W_3+b_3
\end{equation}
Where $\left\{W_1,b_1\right\}$, $\left\{W_2,b_2\right\}$ and $\left\{W_3,b_3\right\}$ are the learning parameters of generator network, respectively. Similar operations are performed on both the synthetic graph $G_g$ and the real graph $G_r$. Then, we apply a softmax layer on the last output hidden state $\mathcal{D}(T)$ to indicate whether it is normal or anomalous.

\subsubsection{Adversarial Training Process}
During the adversarial training process, the conditional graph generator $\mathcal{T}$ will generate the synthetic graph $G_g =\mathcal{T} \left(G_o\right)$ based on the input noisy graph $G_o$. Subsequently, the discriminator $\mathcal{D}$ will try to distinguish between the real and synthetic samples. 

\textbf{Adversarial constraints.} Through adversarial training, the generator $\mathcal{T}$ aims to minimize the adversarial loss function $\mathcal{L}_{GAN}$. While, the discriminator $\mathcal{D}$ aims to maximize the adversarial loss function $\mathcal{L}_{GAN}$. The adversarial loss function is defined as:

\begin{equation}
\label{eq:f10}
\begin{aligned}
\mathcal{L}_{GAN} = min_{\mathcal{T}}max_{\mathcal{D}}\mathbb{E} _{G_r\sim  P(G_r)}[log\mathcal{D}(G_r|C_r^{*})]
\\
+ \mathbb{E}_{G_o\sim P(G_o)}[log(1-\mathcal{D}(\mathcal{T} (G_o)|C_g^{*}))]
\end{aligned}
\end{equation}

\textbf{Distance Constraints.} To further optimize the generator, so that it can not only generate samples that deceive the discriminator, but also ensure that the generated sample distribution close to the real sample distribution, we introduce the reconstruction loss. Here, Mean Absolute Error(MAE) is chosen as its resilience to outliers and higher robustness. By introducing $L_{1}$ distance loss, we can ensure that the generated node properties are closely align with the real ones. The reconstruction loss $\mathcal{L}_{1}$ is defined as:
\begin{equation}
\label{eq:f6}
\mathcal{L}_{1}=\mathbb{E}_{G_r\sim  P(G_r),G_o\sim P(G_o)} [\left \| X_r-\mathcal{T}(X_{o})  \right \|_{1} ],
\end{equation}

To further encourage the generator to generate synthetic graphs that closely resemble the real data, we do not only rely on the reconstruction loss between the synthetic and real data. Instead, we also try to introduce a latent space constraint to minimize the distance between generated embeddings in the high-dimensional latent space. Thus, we define the  between real and synthetic data via $\mathcal{L}_{2}$ loss function, which is defined as:

\begin{equation}
\label{eq:f7}
\begin{aligned}
\mathcal{L}_{2}=\mathbb{E}_{G_r\sim  P(G_r),G_o\sim P(G_o)} [\left \| \mathcal{D}(X_{r})-\mathcal{D}(\mathcal{T}(X_{o}))  \right \|_{2}^{2} ]
\end{aligned}
\end{equation}
By integrating the embedding distance loss $\mathcal{L}_{2}$ into the reconstruction loss $\mathcal{L}_{1}$, we can get the distance constraint $\mathcal{L}_{\mathcal{T}}$ for the generator:
\begin{equation}
\label{eq:f8}
\mathcal{L}_{\mathcal{T}}=\mathcal{L}_{1}+\mathcal{L}_{2}.
\end{equation}

Finally, the overall loss function is defined as:
\begin{equation}
\label{eq:all_loss}
\mathcal{L}_{loss} = \lambda _{1}\mathcal{L}_{GAN}+ \lambda _{2}\mathcal{L}_{\mathcal{T}}
\end{equation}
Here, $\lambda _{1},\lambda _{2},\lambda _{3}$ are the regularization parameter to balance the three losses. The training process is based on the architecture of Wasserstein GAN (\cite{engelmann2021conditional}), which basically solves the problems of slow convergence and collapse mode of GAN by weight clipping. The details are shown in the Algorithm \ref{alg:abnormal}.
\floatname{algorithm}{Algorithm}
\begin{algorithm}[hptb] 
\footnotesize
	\renewcommand{\algorithmicrequire}{\textbf{Input:}}
	\renewcommand{\algorithmicensure}{\textbf{Output:}}
	\caption{Conditional Graph Generative Model}  
	\label{alg:abnormal}
	\begin{algorithmic}[1] 
		\Require
  Matrix noise $A_o$, Label noise $C_o$, Feature noise $X_o$, Graph snapshot dataset $M$, Conditional graph generator $\mathcal{T}$, Discriminator 
 $\mathcal{D}$, Real data $G_o$, Graph numbers that needs to be generated $I$. 

		\Ensure  Model parameter $\theta$, Synthetic data $G_g$. 
   \State  Initialize model parameters   $\theta$.
   \For{each $i=1:I$}
  \For{each $e=1:epoch$}
            \State ${X}_{o}^{*}= GCN({A_o},{X_o})$
            \State $Z\gets $ Eq.\ref{eq:f3};$C_o^{*}\gets$ Eq.\ref{eq:cons}
            \State ${\boldmath{X_g}}={Z}+C_o^{*}$
 \State $\mathcal{L}_{GAN} \gets$ Eq.\ref{eq:f10}
 \State $\mathcal{L}_{1}=\mathbb{E}_{G_r\sim  P(G_r),G_o\sim P(G_o)} [\left \| X_r-\mathcal{T}(X_{o})  \right \|_{1} ] $
 \State $\mathcal{L}_{2} = \mathbb{E}_{G_r\sim  P(G_r),G_o\sim P(G_o)} [\left \| \mathcal{D}(X_{r})-\mathcal{D}(\mathcal{T}(X_{o}))  \right \|_{2}^{2} ] $
\State $\theta = \theta - \eta \bigtriangledown _{\theta }\left \{ \mathcal{L}_{GAN},\mathcal{L}_{1},\mathcal{L}_{2} \right \} $
\EndFor
\State${\boldmath{A^i_g}}=downsampling\left(\boldmath{A^i_o}\right)$
\State Collect generation $G^i_g\left(\boldmath{A^i_g},\boldmath{X^i_g},\boldmath{C^i_g}\right)$ for the $i$ round.
\EndFor

\Return{$\theta$}, Synthetic data $G_g$.

\end{algorithmic}
\end{algorithm}

\subsection{GNN-Based Node Anomaly Detection}
In this section, we briefly introduced the process of anomaly detection using graph neural networks~(GNNs). The model takes the feature matrix $X\in{\mathbb{R}^{N\times F}}$ and the adjacency matrix $A\in{\mathbb{R}^{N\times N}}$ of each graph snapshot as input. The topological features of the nodes in each graph snapshot are extracted by a graph neural network and output as embedding vectors $C_i$, 
\begin{equation}
\hat{C} _{i}=\text{GNN}(X_{i}, A_{i}).
 \label{equ:gnn}
\end{equation}
    
As shown in Algorithm \ref{alg:model}, the graph embedding vector $\{\hat{C} _{i}|i=1,2,\ldots,I\}$, where $I$ is the number of graph snapshots, is obtained from GNN-based learning model. Then the cross-entropy loss function is used to train the anomaly detection model. Eventually, by minimizing the cross-entropy loss function during training, the model learns to identify the behavior of nodes, and have the ability to detect different categories of anomalies.

\floatname{algorithm}{Algorithm}
\begin{algorithm}[hptb] 
\footnotesize
	\renewcommand{\algorithmicrequire}{\textbf{Input:}}
	\renewcommand{\algorithmicensure}{\textbf{Output:}}
	\caption{GNN-Based Node Anomaly Detection}  
	\label{alg:model}
	\begin{algorithmic}[1] 
		\Require Adjacency matrix $\left\{A_1,A_2,\ldots,A_{I}\right\}$, Feature matrix $\left\{X_1,X_2,\ldots,X_{I}\right\}$, Label matrix $\{C_1,C_2,\ldots,C_{I}\}$, Iteration $epoch$, Number of graph $I$.
		\Ensure  model parameter $\theta$.
   \State  Initialisation parameters $\theta$.
        \For{each $e=1:epoch$}
        \For{each $i=1:I$}
           \State $\hat{C} _{i} \gets \text{GNN}(X_i, A_i)$
           \State $\mathcal{L}= {\rm{cross\_entropy}}\left(C_i,\hat{C} _{i}\right)$
            \EndFor    
	\State Updating model parameters $\theta$ based on gradient back propagation
           
      \State \Return{$\theta$}
    \EndFor
\end{algorithmic}
\end{algorithm}

\section{Experiment}\label{sec:experiment}
\subsection{Experimental Settings}
\subsubsection{Datasets}

To generate effective graph snapshot for the task of anomaly detection on nodes, the experimental dataset must contain features representing network topology information such as IP addresses and ports. It also needs to contain multiple attribute features that can identify traffic classes. Finally, the temporal distribution of the attacks should be relatively uniform. Considering the above requirements, we select two public datasets, namely UNSW-NB15 and CICIDS-2017 to evaluate the effectiveness of the proposed method. Both datasets contain newer data.

\par
The UNSW-NB15 (\cite{basati2022pdae}) dataset is now one of the most commonly used benchmark datasets in the field of cyber security. The dataset contains 9 types of network attacks which are Fuzzers, Analysis, Backdoor, DoS, Exploits, Generic, Reconnaissance and Worms. The Category proportion are highly imbalanced, such as Worms account for only 0.007\% of the total.

\par
The CICIDS-2017 (\cite{chapaneri2022enhanced}) is a Netflow-based simulation dataset with $78$ features. 
The dataset covers a variety of attack types including Web Attack, Brute force, DoS, DDoS, Infiltration, Heart-bleed, Bot and Scan. For the convenience of time sampling in generating TDG, we selected samples from only six typical categories.
\subsubsection{Evaluation metrics}
The five metrics of Accuracy, Recall, False alarm rate (FAR), Precision and F1-Score have been widely used to evaluate the performance of classification models.\par
In order to evaluate the correlation between the generated data and the real data, we choose three correlation measures.
Where Wasserstein Distance (\cite{rubner2000earth}) can cope with the problem that JS dispersion does not measure the distance between two distributions that are not overlapping. KStest (\cite{strickland2023drl}) assesses the similarity of continuous features, and KSTest uses the Two-sample Kolmogorov–Smirnov test and the empirical Cumulative Distributed Function(CDF) to compare columns with continuous values to their distributions. Finally, we also calculated the Maximum Mean Discrepancy(MMD) (\cite{gretton2012kernel}) between the two data distributions.

\subsubsection{Baselines}
We compare the ability of different generation models to generate synthetic graph snapshot. These are two tabular data generation models CTGAN (\cite{xu2019modeling}), TableGAN (\cite{park2018data}), and two graph data generative models GraphRNN (\cite{you2018graphrnn}) , GraphSGAN (\cite{ding2018semi}).
\begin{itemize}
	\item CTGAN: The method addresses the challenges of synthetic tabular data generation for pattern normalization and data imbalance issues.
	\item TableGAN: It is a synthetic data generation technique which has been implemented using a deep learning model based on Generative Adversarial Network architecture. 
	\item GraphRNN: GraphRNN is an autoregressive generative model is built on Graphs under the same node ordering are represented as sequences.
        \item GraphSGAN: The GraphSGAN framework addresses the problem of having a graph consisting of a small set of labeled nodes and a set of unlabeled nodes, and how to learn a model that can predict the labeling of the unlabeled nodes. 
\end{itemize}

\subsection{Binary Anomaly Detection Evaluation}
In this section, we evaluate the effectiveness of CGGM compared
with other generation methods on UNSW-NB15 and CICIDS-2017 to verify whether CGGM works as expected. Most generation models do not support end-to-end multi-class generation. Therefore, we chose to conduct experiments using real labels. As observed, the recall of CGGM data on the UNSW-NB15 dataset is 0.98, indicating an excellent model fit. Similarly, it performs remarkably well on the CICIDS-2017 dataset, significantly outperforming the subpar CTGAN. It is evident that models trained with synthetic data generated by CGGM achieve the best performance, while data generated by other models exhibit more unstable training results.
\begin{table*}[htbp]
\begin{center}
\renewcommand{\tablename}{Table}
\caption{Classification performance comparisons on UNSW-NB15, CICIDS-2017 datasets.}   
\footnotesize
\begin{tabular*}{\textwidth}{@{\extracolsep{\fill}}cccccc@{}}
      \toprule
{$\rm Data$}&{$\rm Method$}&{$\rm Accuracy$}&{$\rm Recall$}&{$\rm Precision$}&{$\rm F1-score$}\\
\midrule
\multirow{4}{*}{{$\rm UNSW-NB15$}}{}&{$\rm TableGAN$~\cite{park2018data}}&{$\rm 0.65$$\pm$0.13}&{$\rm 0.93$$\pm$0.03}&{$\rm 0.61$$\pm$0.12}&{$\rm 
 0.74$$\pm$0.09}\\
&{$\rm CTGAN$~\cite{xu2019modeling}}&{$\rm 0.91$$\pm$0.02}&{$\rm 0.48$$\pm$0.04}&{$\rm 0.38$$\pm$0.02}&{$\rm 0.43$$\pm$0.05}\\
&{$\rm GraphSGAN$~\cite{ding2018semi}}&{$\rm 0.90$$\pm$0.04}&{$\rm 0.48$$\pm$0.08}&{$\rm 0.38$$\pm$0.08}&{$\rm 0.42$$\pm$0.12}\\
&{$\rm GraphRNN$~\cite{you2018graphrnn}}&{$\rm 0.20$$\pm$0.08}&{$\rm 0.18$$\pm$0.03}&{$\rm 0.16$$\pm$0.04}&{$\rm 0.15$$\pm$0.03}\\
&\boldmath\textbf{{$\rm CGGM(ours)$}}&\boldmath\textbf{0.98$\pm$0.01}&\boldmath\textbf{0.98$\pm$0.01}&\boldmath\textbf{0.98$\pm$0.01}&\boldmath\textbf{0.98$\pm$0.01}\\
\\
\multirow{4}{*}{{$\rm CICIDS-2017$}}&{$\rm TableGAN$~\cite{park2018data}}&{$\rm 0.67$$\pm$0.15}&{$\rm 0.98$$\pm$0.02}&{$\rm 0.67$$\pm$0.17}&{$\rm 0.79$$\pm$0.10}\\
&{{$\rm CTGAN$~\cite{xu2019modeling}}}&{$\rm 0.88$$\pm$0.05}&{$\rm 0.95$$\pm$0.10}&{$\rm 0.88$$\pm$0.05}&{$\rm 0.92$$\pm$0.02}\\
&{$\rm GraphSGAN$~\cite{ding2018semi}}&{$\rm 0.90$$\pm$0.06}&{$\rm 0.07$$\pm$0.24}&{$\rm 0.13$$\pm$0.27}&{$\rm 0.11$$\pm$0.12}\\
&{$\rm GraphRNN$~\cite{you2018graphrnn}}&{$\rm 0.80$$\pm$0.06}&{$\rm 0.65$$\pm$0.09}&{$\rm 0.74$$\pm$0.21}&{$\rm 0.67$$\pm$0.13}\\
&\boldmath\textbf{{$\rm CGGM(ours)$}}&\boldmath\textbf{0.97$\pm$0.02}&\boldmath\textbf{0.98$\pm$0.01}&\boldmath\textbf{0.96$\pm$0.02}&\boldmath\textbf{0.97$\pm$0.02}\\
\bottomrule
\end{tabular*}
\label{tab:third}
\end{center}
\end{table*}

\begin{figure*}[htbp]
\centering  
\subfloat[CGGM]{\label{fig:subfig1}\includegraphics[width=0.2\linewidth]{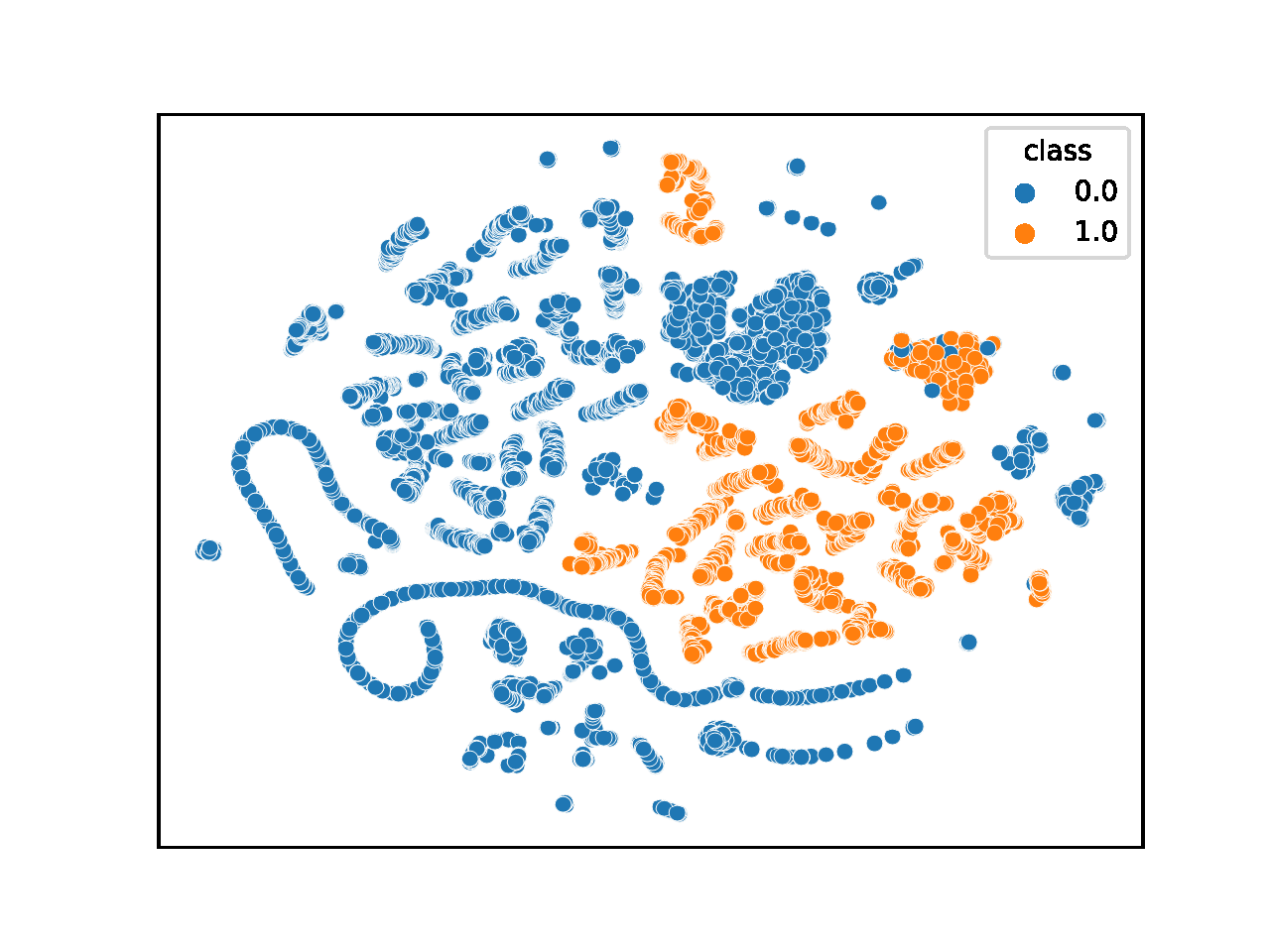}}
\subfloat[TableGAN]{\label{fig:subfig2}\includegraphics[width=0.2\linewidth]{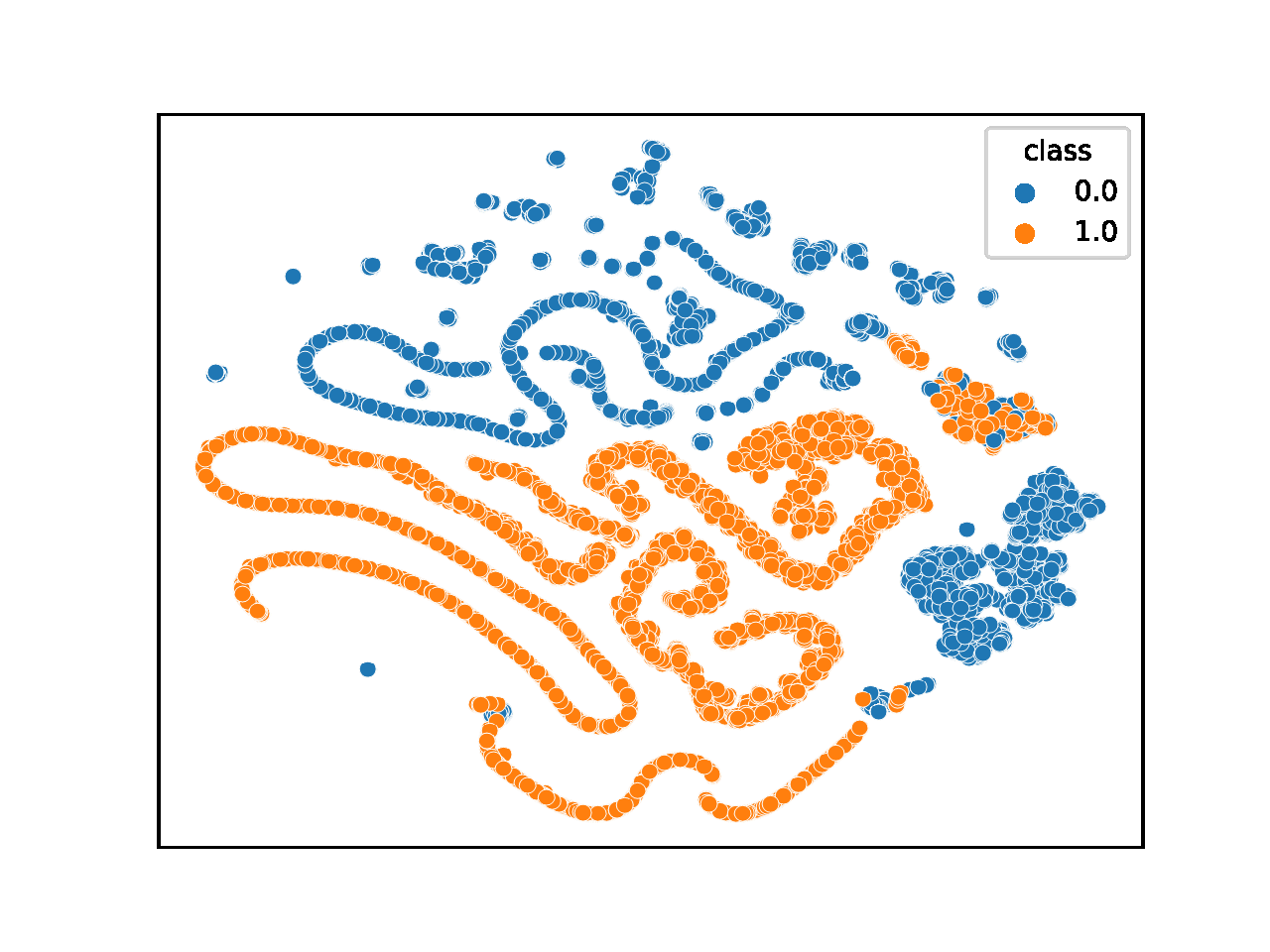}}
\subfloat[CTGAN]{\label{fig:subfig2}\includegraphics[width=0.2\linewidth]{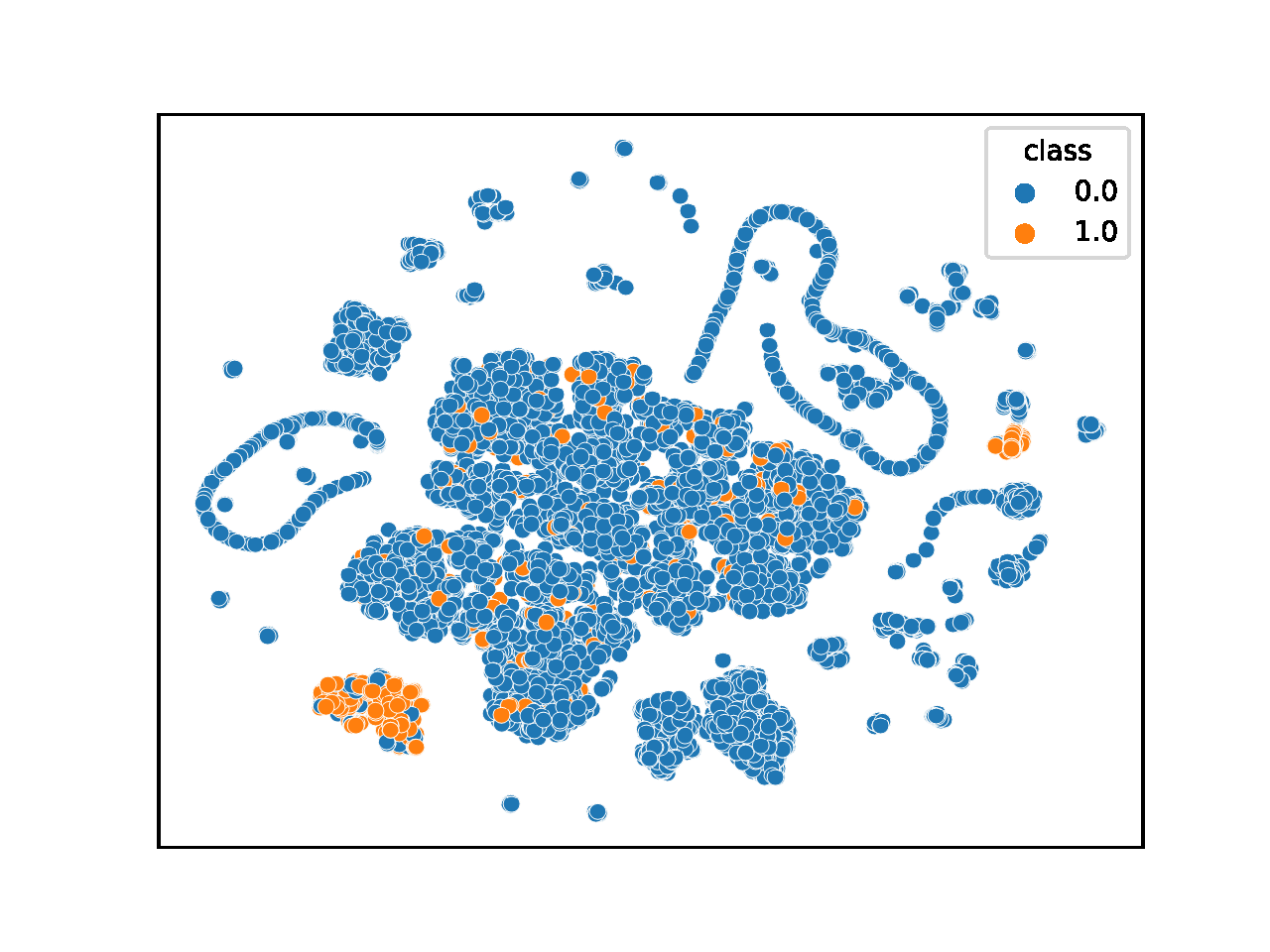}}
\subfloat[GraphRNN]{\label{fig:subfig2}\includegraphics[width=0.2\linewidth]{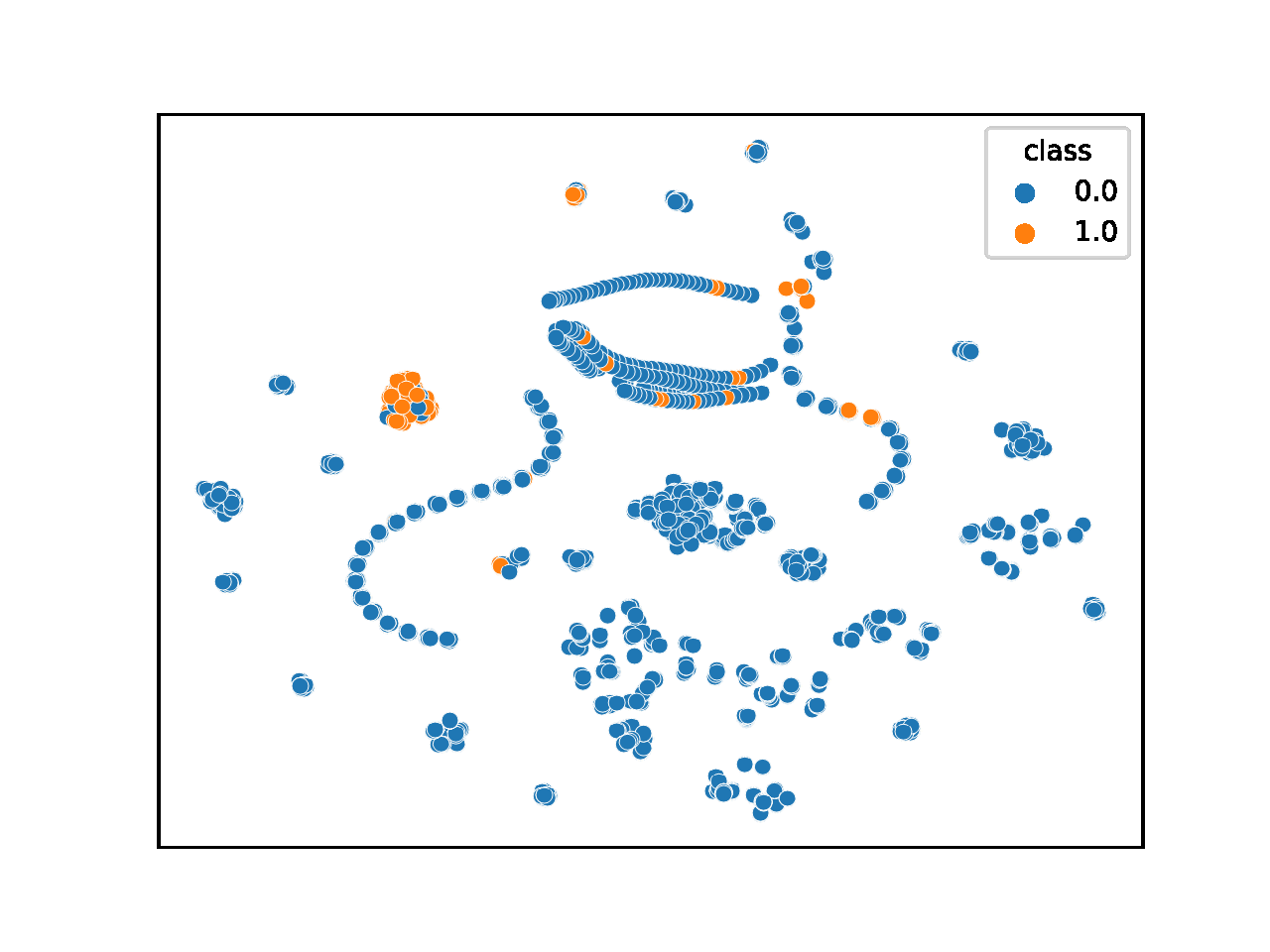}}
\subfloat[GraphSGAN]{\label{fig:subfig2}\includegraphics[width=0.2\linewidth]{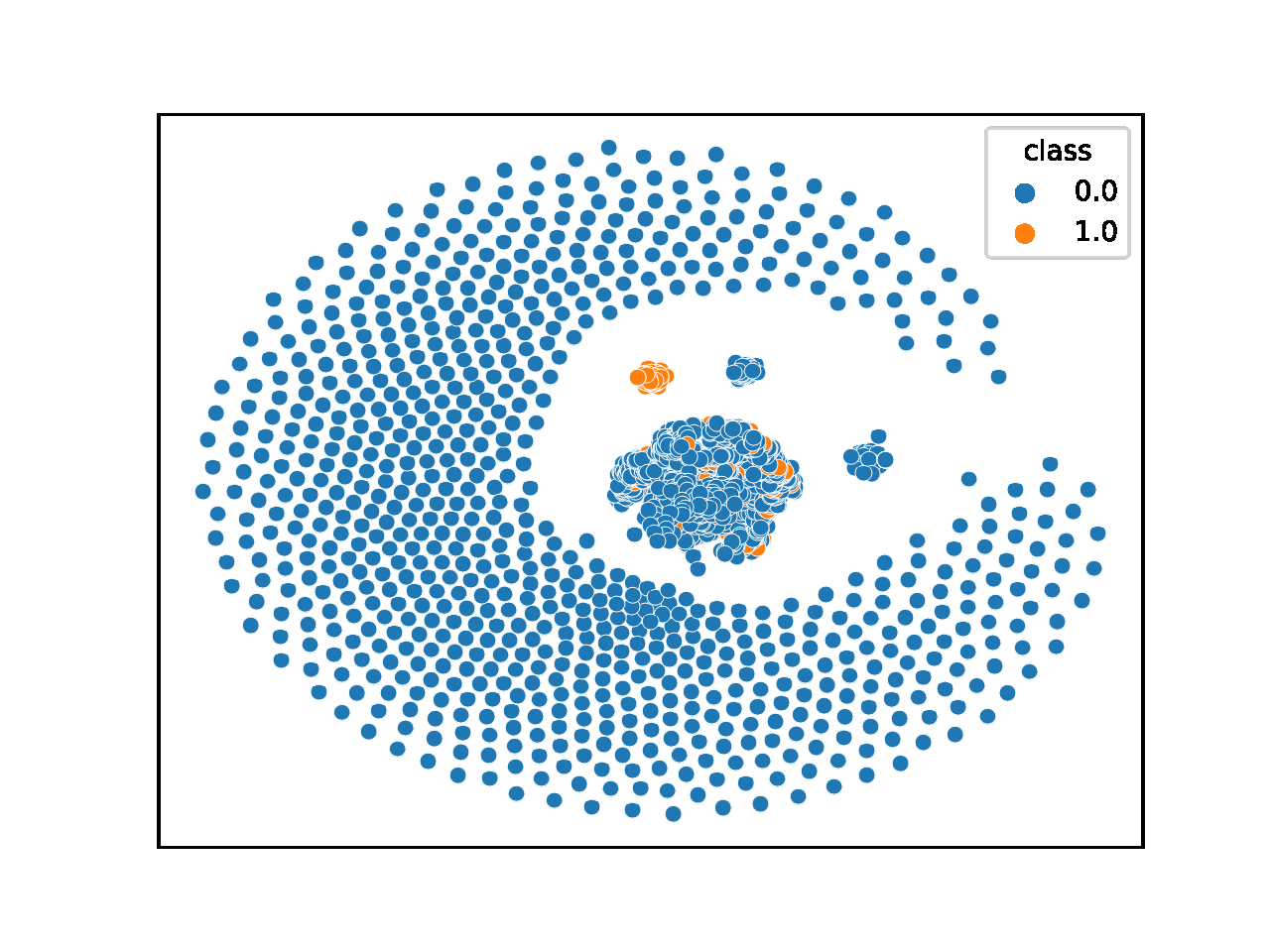}}
	\caption{T-SNE visualisation of synthetic data features with different generation methods.}
 \label{fig:sne}
\end{figure*}

\begin{figure}[htbp]
\centering  
\subfloat[UNSW-NB15]{\label{fig:subfig1}\includegraphics[width=0.4\linewidth]{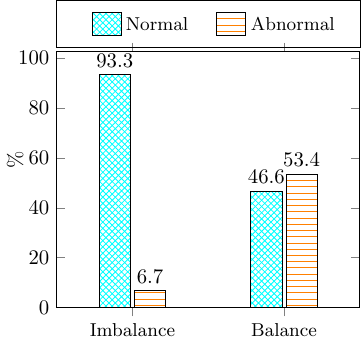}}
		\subfloat[CICIDS-2017]{\label{fig:subfig2}\includegraphics[width=0.4\linewidth]{Fig3_1.pdf}}
	 \caption{Label category balance result.}
 \label{fig:train}
\end{figure}

We have provided a detailed explanation of the process and results pertaining to category balancing on the UNSW-NB15 and CICIDS-2017 datasets. As shown in Fig. \ref{fig:train}, there is a noticeable disparity in category proportions within the UNSW-NB15 dataset. The initial distribution between normal and abnormal classes was 93.3\% and 6.7\%, respectively. Our efforts led to a transformation of these proportions into a more balanced 50\% for both categories. In contrast, the category balance in the CICIDS-2017 dataset remained relatively even. Through the generation of data for minority classes, we once again increased the proportion of these minority categories to 50.5\%.

In Fig. \ref{fig:sne}, we present the visualizations of features for both the real and synthetic data in the UNSW-NB15 dataset. The synthetic data generated by CGGM exhibits a distribution that closely resembles the original data, with clearly defined classification boundaries for labeled categories. It is noteworthy that while methods like TableGAN and CTGAN can measure the similarity of distributions in feature dimensions, they fall short in capturing the graph topological associations.

We further list the correlation scores for binary evaluation on UNSW-NB15, which are 0.0818, 0.6181, and 0.4246, respectively. The detailed experimental results are shown in Tab. \ref{tab:second}. It can be intuitively seen that CGGM consistently outperforms all other baseline models. Compared with the baselines, CGGM introduced more constraints in the training process, which provides more necessary feature information and label signals for the required patterns of the synthetic graph.

\begin{table}[htbp]
\footnotesize
\centering
        \caption{Distribution discrepancy between synthetic and real data with Wasserstein, MMD, and KSTest metircs.} 
        \begin{tabular*}{0.475\textwidth}[htbp]{@{\extracolsep{\fill}}cccc@{\extracolsep{\fill}}}
      
        \toprule
{$\rm Method$}&{$\rm Wasserstein$}&{$\rm MMD$}&{$\rm KSTest$}\\
\toprule
{$\rm TableGAN$~\cite{park2018data}}&{$\rm 0.3404$}&{$\rm 0.7479$}&{$\rm 0.8227$}\\
{$\rm CT$-$\rm{GAN}$~\cite{xu2019modeling}}&{$\rm 0.0935$}&{$\rm 0.7375$}&{$\rm 0.7048$}\\
{$\rm GraphRNN$~\cite{you2018graphrnn}}&{$\rm 0.1034$}&{$\rm 2.4527$}&{$\rm 0.5175$}\\
{$\rm GraphSGAN$~\cite{ding2018semi}}&{$\rm -
$}&{$\rm 1.7614$}&{$\rm 0.4679$}\\
 \boldmath\textbf{$\rm CGGM(ours) $}& \boldmath\textbf{0.0818}& \boldmath\textbf{0.6181}&\boldmath\textbf{0.4246}\\
        \bottomrule
        \end{tabular*}
        \label{tab:second}
\end{table}

\subsection{Multi-class Anomaly Detection Evaluation}

\begin{table*}[htbp]
\footnotesize
\caption{The statistics of the different datasets on categories proportion.}
\centering
\begin{tabular*}{\textwidth}{@{\extracolsep{\fill}}cccccccc@{\extracolsep{\fill}}}
\toprule
\multirow{2}{*}{{$\rm Data$}}&\multirow{2}{*}{{$\rm Category$}}&\multicolumn{2}{c}{{$\rm Original$}} &\multicolumn{2}{c}{{$\rm Real Graph$}} &\multicolumn{2}{c}{{$\rm Generated Graph$}} \\
\cmidrule(lr){3-4}\cmidrule(lr){5-6}\cmidrule(lr){7-8}&&{$\rm Number$}&{Pct.}&{$\rm Number$}&{Pct.}&{$\rm Number$}&{Pct.}\\
\midrule
\multirow{9}{*}{{$\rm UNSW-NB15$}}&{$\rm Normal$}&{$\rm 2218761$}&{$\rm 87.3\%$}&{$\rm 4957$}&{$\rm 71.7\%$}&{$\rm 561$}&{$\rm 8.1\%$}\\
&{$\rm Fuzzers$}&{$\rm 24246$}&{$\rm 0.95\%$}&{$\rm 862$}&{$\rm 12.5\%$}&{$\rm 840$}&{$\rm 12.1\%$}\\
&{$\rm Analysis$}&{$\rm 2677$}&{$\rm 0.11\%$}&{$\rm 246$}&{$\rm 3.5\%$}&{$\rm 560$}&{$\rm 8.1\%$}\\
&{$\rm Backdoors$}&{$\rm 2329$}&{$\rm 0.10\%$}&{$\rm 58$}&{$\rm 0.8\%$}&{$\rm 560$}&{$\rm 8.1\%$}\\
&{$\rm DoS$}&{$\rm 16353$}&{$\rm 0.64\%$}& {$\rm 215$}&{$\rm 3.1\%$}&{$\rm 280$}&{$\rm 4.0\%$}\\
&{$\rm Exploits$}&{$\rm 44525$}&{$\rm 1.75\%$}&{$\rm 173$}&{$\rm 2.5\%$}&{$\rm 840$}&{$\rm 12.1\%$}\\
&{$\rm Generic$}&{$\rm 215481$}&{$\rm 8.40\%$}&{$\rm 165$}&{$\rm 2.3\%$}&{$\rm 840$}&{$\rm 12.1\%$}\\
&{$\rm Reconnaissance$}&{$\rm 13987$}&{$\rm 0.56\%$}&{$\rm 212$}&{$\rm 3.0\%$}&{$\rm 560$}&{$\rm 8.1\%$}\\
&{$\rm Shellcode$}&{$\rm 1511$}&{$\rm 0.057\%$}&{$\rm 13$}&{$\rm 0.1\%$}&{$\rm 1400$}&{$\rm 20.3\%$}\\
&{$\rm Worms$}&{$\rm 174$}&{$\rm 0.007\%$}&{$\rm 8$}&{$\rm 0.1\%$}&{$\rm 420$}&{$\rm 6.1\%$}\\
\bottomrule
\end{tabular*}
\label{tab:Time}
\end{table*}

Tab. \ref{tab:Time} shows the structure of the UNSW-NB15, especially the  proportion of all attack classes. It can be intuitively seen that the original UNSW-NB15 dataset is highly imbalanced, with even greater variations between the individual attack classes, as shown in Tab. \ref{tab:Time}, where the normal class has a high proportion of 87.3\% the least Warms attack type is only 0.007\% of the total, which can greatly affect the learning performance of the model. The resulting graph snapshot also has a very imbalanced class proportion. As shown in Tab. \ref{tab:Time}, the proportion of normal categories is still as high as 71.7\%, and a few attack categories such as Shellcode and Warms only account for 0.1\% of the total. We try to make a balance of the multi-target classification data by CGGM to consider all attack types separately and reshape their distribution. The data generated by CGGM has more reasonable category proportions, and most of the category proportions are balanced to about 10\%. At the same time, the number of samples for each attack family is homogenised to improve the classification results for specific attack categories.

We compared the optimal results achieved by each model with identical configurations. To validate the effectiveness and applicability of synthetic data, we trained the models using synthetic data and tested them using real data. The experimental results are presented in Tab. \ref{tab:mutir}. We can observe that experiments based on different datasets consistently show higher performance of the GCN model compared to the GraphSAGE model, indicating that the GCN model is more suitable for node anomaly detection tasks. Furthermore, it is evident that training on balanced datasets significantly improves the model's classification performance.

\begin{table*}[htbp]
\begin{center}
\footnotesize
\caption{Classification performance of multi-category results with different backbones.}   
\centering
\begin{tabular*}{\linewidth}{@{\extracolsep{\fill}}cccccccc}
\toprule
{$\rm Data$}&{$\rm Model$}&{$\rm Datatype$}&{$\rm Accuracy$}&{$\rm Recall$}&{$\rm Precision$}&{$\rm F1-score$}\\
\midrule
\multirow{4}{*}{{$\rm UNSW-NB15$}}
&\multirow{2}{*}{{$\rm GCN$}}&{$\rm imbalance$}&{$\rm 0.73$$\pm$0.21}&{$\rm 0.14$$\pm$0.10}&{$\rm 0.14$$\pm$0.10}&{$\rm 0.12$$\pm$0.08}\\
&&{$\rm banlance$}&\boldmath\textbf{0.95$\pm$0.03}&\boldmath\textbf{0.96$\pm$0.04}&\boldmath\textbf{0.96$\pm$0.04}&\boldmath\textbf{0.96$\pm$0.02}\\
&\multirow{2}{*}{{$\rm GraphSAGE$}}&{$\rm imbalance$}&{$\rm 0.72$$\pm$0.13}&{$\rm 0.10$$\pm$0.10}&{$\rm 0.07$$\pm$0.10}&{$\rm 0.08$$\pm$0.08}\\
&&{$\rm banlance$}&\boldmath\textbf{0.98$\pm$0.01}&\boldmath\textbf{0.98$\pm$0.01}&\boldmath\textbf{0.98$\pm$0.01}&\boldmath\textbf{0.98$\pm$0.02}\\
\\
\multirow{4}{*}{{$\rm CICIDS-2017$}}
&\multirow{2}{*}{{$\rm GCN$}}&{$\rm imbalance$}&{$\rm 0.71$$\pm$0.14}&{$\rm 0.17$$\pm$0.08}&{$\rm 0.26$$\pm$0.14}&{$\rm 0.14$$\pm$0.09}\\
&&{$\rm banlance$}&\boldmath\textbf{0.93$\pm$0.03}&\boldmath\textbf{0.96$\pm$0.01}&\boldmath\textbf{0.96$\pm$0.01}&\boldmath\textbf{0.96$\pm$0.02}\\
&\multirow{2}{*}{{$\rm GraphSAGE$}}&{$\rm imbalance$}&{$\rm 0.71$$\pm$0.21}&{$\rm 0.17$$\pm$0.12}&{$\rm 0.29$$\pm$0.20}&{$\rm 0.17$$\pm$0.17}\\
&&{$\rm banlance$}&\boldmath\textbf{0.98$\pm$0.01}&\boldmath\textbf{0.98$\pm$0.01}&\boldmath\textbf{0.98$\pm$0.01}&\boldmath\textbf{0.98$\pm$0.01}\\
\bottomrule
\end{tabular*}
\label{tab:mutir}
\end{center}
\end{table*}

\begin{figure*}[htbp]

\centering  
\subfloat[Classification performance with different data categories on UNSW-NB15.]{\label{fig:subfig1}\includegraphics[width=0.6\linewidth]{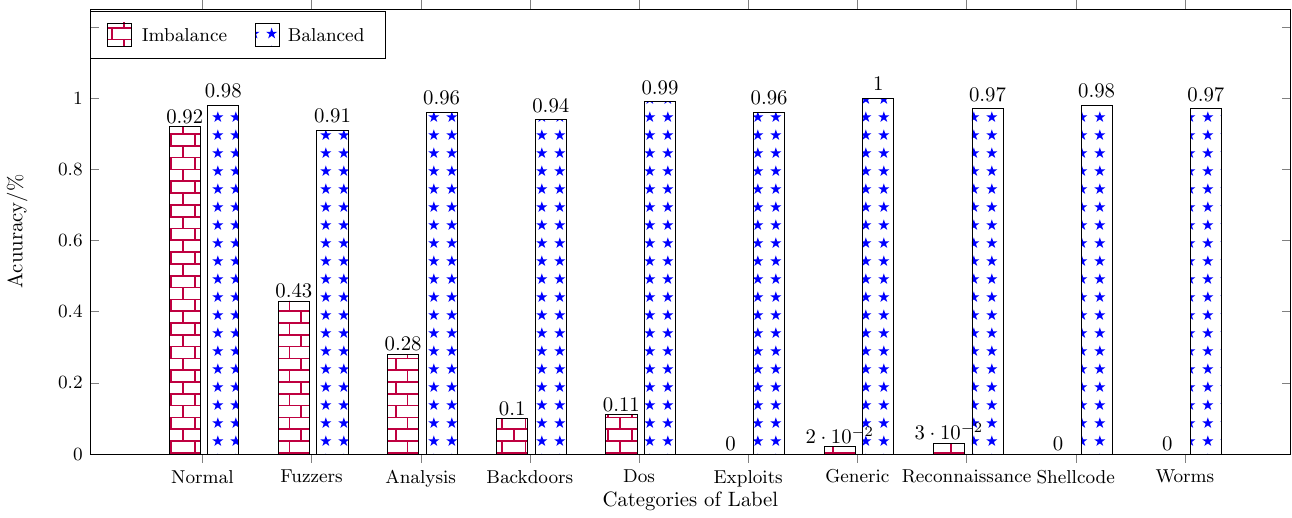}}
\subfloat[Confusion matrix on balanced UNSW-NB15.]{\label{fig:subfig2}\includegraphics[width=0.3\linewidth]{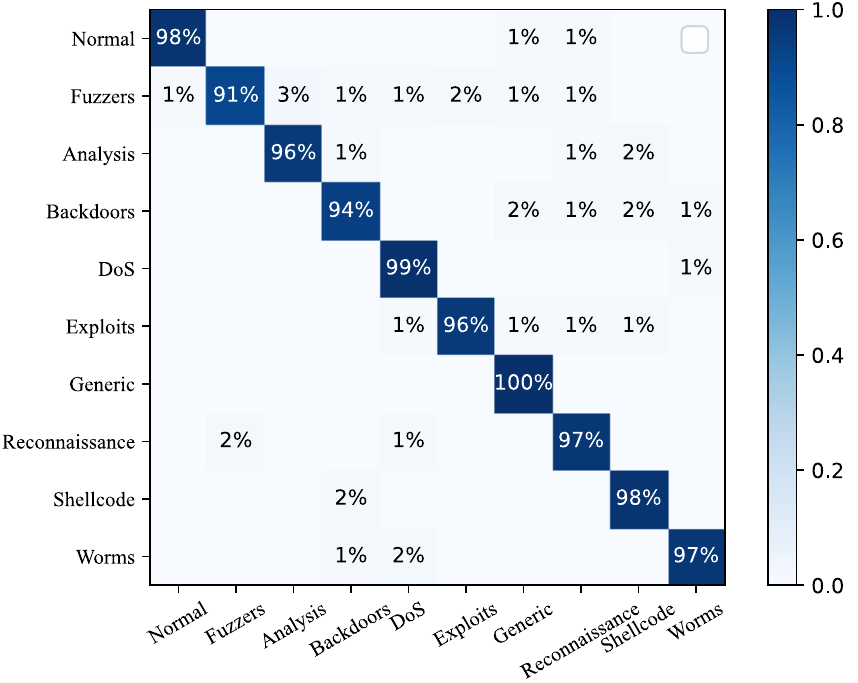}} \\

\subfloat[Classification performance with different data categories on CICIDS-2017.]{\label{fig:subfig1}\includegraphics[width=0.6\linewidth]{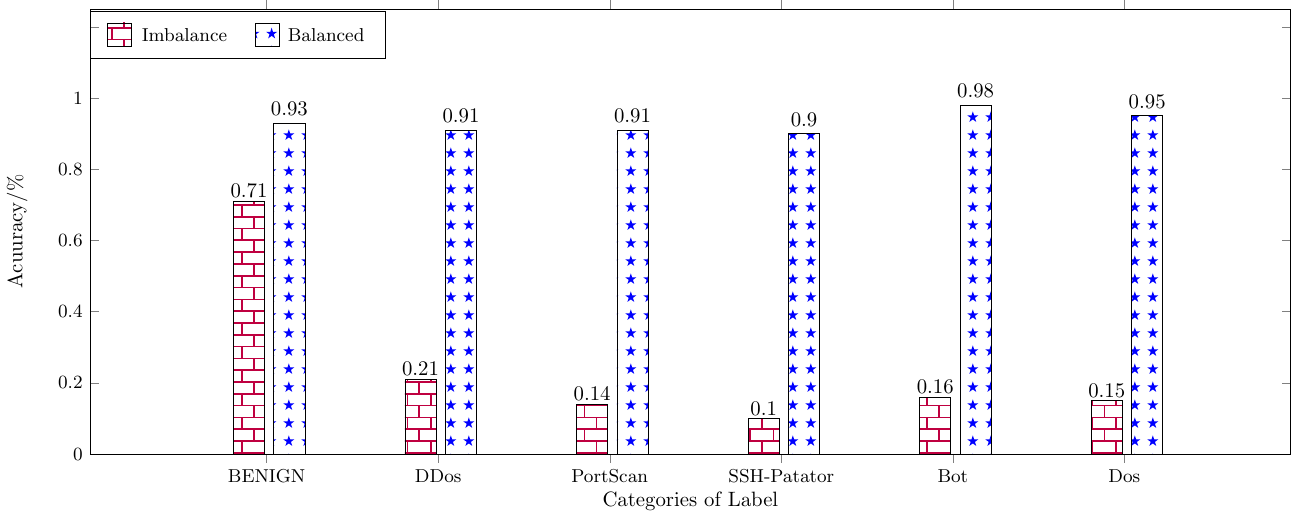}}
\subfloat[Confusion matrix on balanced CICIDS-2017.]{\label{fig:subfig2}\includegraphics[width=0.3\linewidth]{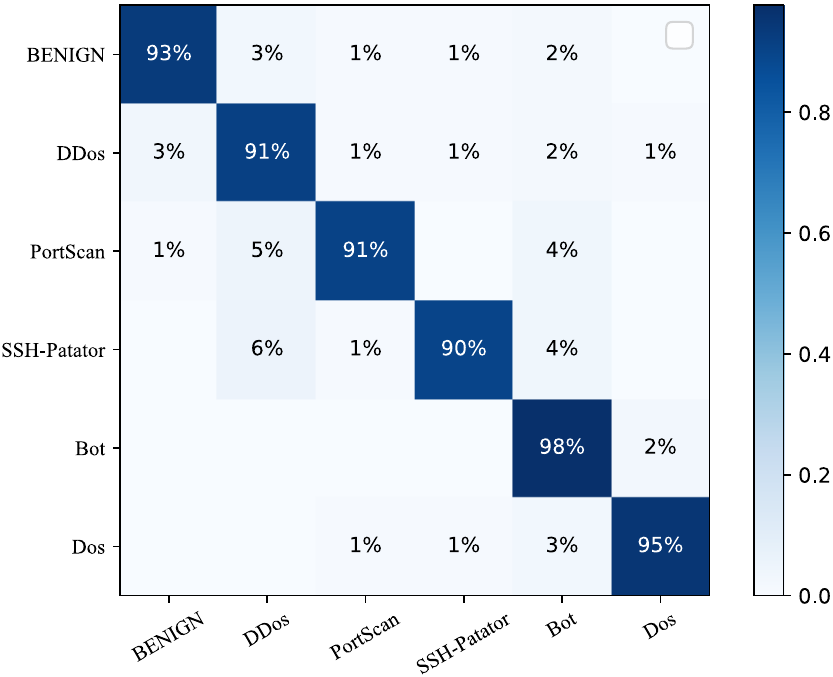}} \\
	\caption{Multi-classification performance with balanced and imbalanced datasets.} 
 \label{fig:mutiac}
\end{figure*}

Fig. \ref{fig:mutiac} provides a detailed illustration of the classification accuracy for each imbalanced and balanced dataset. It is evident that balancing the data categories significantly enhances the model's performance for certain classes. Notably, the accuracy of Fuzzers and Analysis attacks improves dramatically, rising from 0.43 to 0.91 and 0.28 to 0.96, respectively, in the UNSW-NB15 dataset. Several other categories that previously almost learned no information, demonstrated improvements in accuracy to over 0.97 after balancing. Similar to CICIDS-2017, the learning accuracy of DDoS and PortScan improve from 0.21 to 0.91 and 0.14 to 0.91, respectively. By utilizing a balanced dataset, the model's inclination toward the majority class is considerably diminished, leading to a notable improvement in the classification performance for the minority class. 

Meanwhile, we have visualized the features of multi-category data. We can still observe the expanded multi-class anomalous data, which can be effectively distinguished from the normal classes without disrupting the original data distribution pattern. All the aforementioned findings suggest that utilizing the CGGM model to generate synthetic data for augmenting minority classes is a promising approach. This can enhance the performance of the learning model in identifying underrepresented categories.

\begin{figure}[h!]

\centering  
\subfloat[UNSW-2015.]{\label{fig:subfig1}\includegraphics[width=0.48\linewidth]{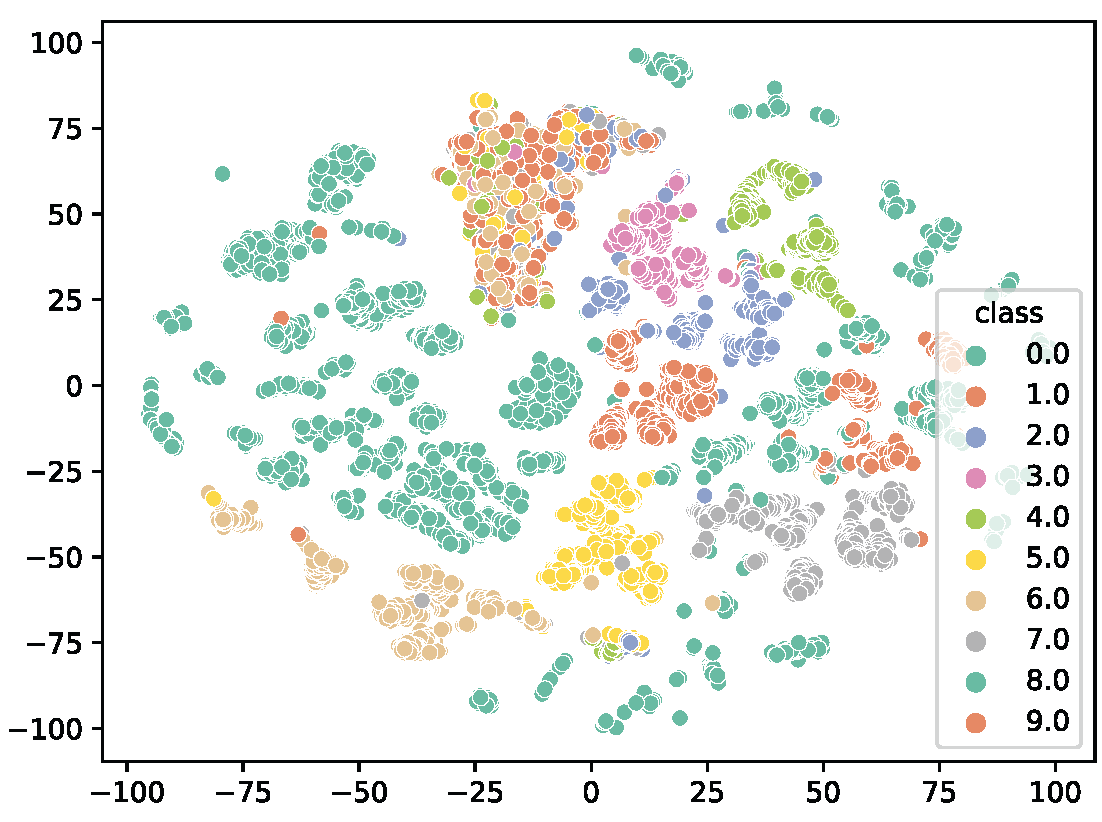}}
\subfloat[CICIDS-2017.]{\label{fig:subfig2}\includegraphics[width=0.48\linewidth]{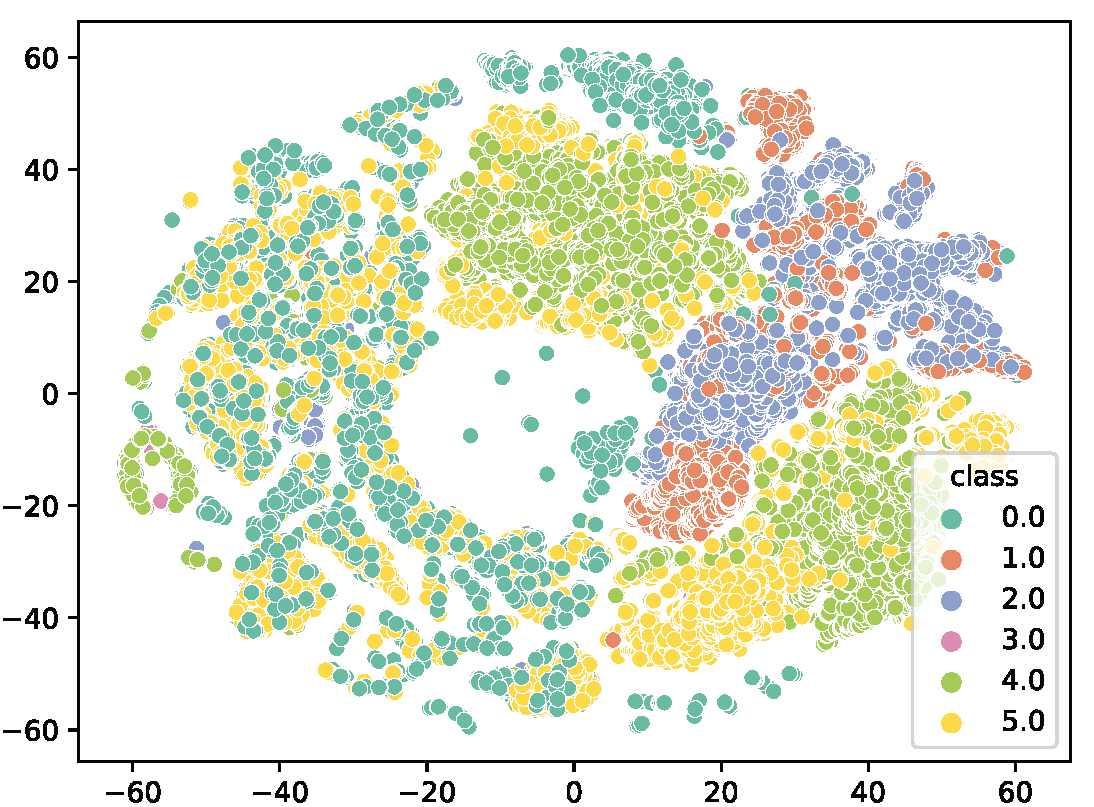}} \\

\caption{T-SNE visualization of multi-category data embeddings on UNSW-NB15 and CICIDS-2017. Each color represents a data category.}
 \label{fig:mutisne}
\end{figure}

\begin{figure}[h!]
\centering  
\subfloat[UNSW-NB15.]{\label{fig:subfig1}\includegraphics[width=0.48\linewidth]{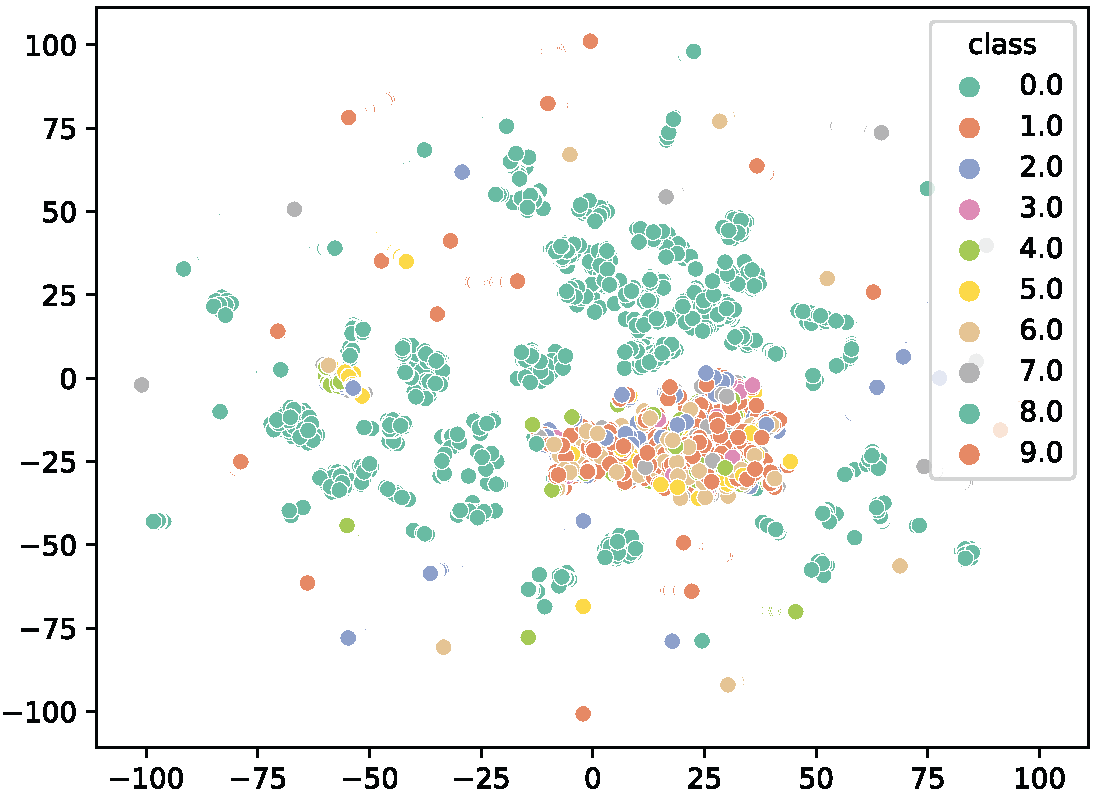}}
\subfloat[CICIDS-2017.]{\label{fig:subfig2}\includegraphics[width=0.48\linewidth]{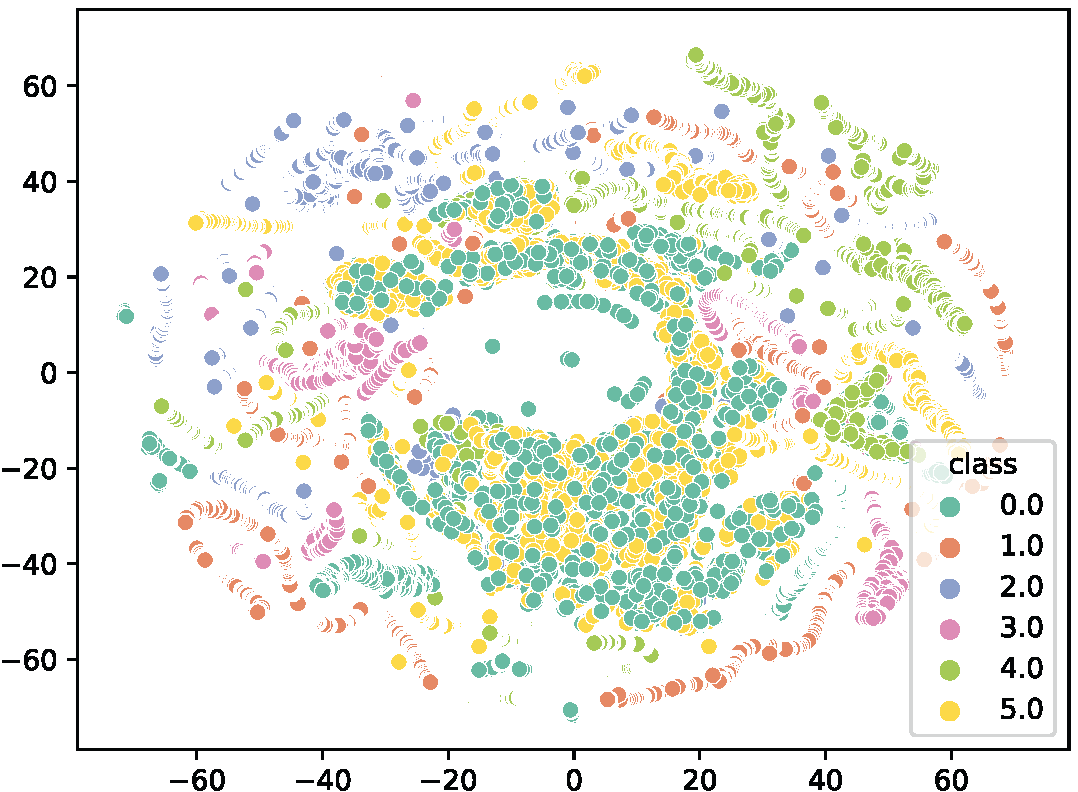}} \\
\caption{T-SNE visualization of data embeddings on UNSW-NB15 and CICIDS-2017. Each color represents a data category.}
 \label{fig:mutifea}
\end{figure}

\subsection{Ablation Study}
\subsubsection{Different Generation Backbones Analysis}
\begin{table*}[htbp]
\footnotesize
\caption{Classification performance with different feature generation backbones.} 
\centering
\begin{tabular*}{\textwidth}{@{\extracolsep{\fill}}cccccccc}
\toprule
{$\rm Data$}&{$\rm Model$}&{$\rm Accuracy$}&{$\rm Recall$}&{$\rm Precision$}&{$\rm F1-score$}\\
\midrule
\multirow{2}{*}{{$\rm UNSW-NB15$}}
&{{$\rm Linear$}}&{$\rm 0.16$$\pm$0.09}&{$\rm 0.10$$\pm$0.08}&{$\rm 0.10$$\pm$0.08}&{$\rm $$0.03$$\pm$0.05}\\
&{$\rm MFG$}&\boldmath\textbf{0.95$\pm$0.04}&\boldmath\textbf{0.96$\pm$0.02}&\boldmath\textbf{0.96$\pm$0.02}&\boldmath\textbf{0.96$\pm$0.03}\\
\\
\multirow{2}{*}{{$\rm CICIDS-2017$}}
&{{$\rm Linear$}}&{$\rm 0.94$$\pm$0.02}&{$\rm 0.96$$\pm$0.02}&{$\rm 0.96$$\pm$0.02}&{$\rm 0.96$$\pm$0.02}\\
&{$\rm MFG$}&\boldmath\textbf{0.98$\pm$0.01}&\boldmath\textbf{0.98$\pm$0.01}&\boldmath\textbf{0.98$\pm$0.01}&\boldmath\textbf{0.98$\pm$0.01}\\
\bottomrule
\end{tabular*}
\label{tab:fea}
\end{table*}

\begin{table*}[htbp]
\footnotesize
\caption{Different graph classifiers performance on UNSW-NB15 and CICIDS-2017.}   
\begin{tabular*}{\textwidth}{@{\extracolsep{\fill}}ccccccccccccc@{}}
      \toprule
   \multirow{2}{*}{{$\rm Model$}}&\multirow{2}{*}{{$\rm Datatype$}}&\multicolumn{4}{c}{{$\rm UNSW-NB15$}}&\multicolumn{4}{c}{{$\rm CICIDS-2017$}}\\

\cmidrule(lr){3-6}\cmidrule(lr){7-10}&&{$\rm Accuracy$}&{$\rm Recall$}&{$\rm Precision$}&{$\rm F1-score$}&{$\rm Accuracy$}&{$\rm Recall$}&{$\rm Precision$}&{$\rm F1-score$}\\
\midrule
\multirow{2}{*}{{$\rm GCN$}}&{$\rm Imbalance$}&{$\rm 0.94$$\pm$0.02}&{$\rm 0.92$$\pm$0.02}&{$\rm 0.80$$\pm$0.08}&{$\rm 0.79$$\pm$0.08}&{$\rm 0.95$$\pm$0.01}&{$\rm 0.93$$\pm$0.02}&{$\rm 0.85$$\pm$0.06}&{$\rm 0.92$$\pm$0.03}\\
&\boldmath\textbf{{$\rm Banlance$}}&\boldmath\textbf{0.98$\pm$0.01}&\boldmath\textbf{0.98$\pm$0.01}&\boldmath\textbf{0.98$\pm$0.01}&\boldmath\textbf{0.98$\pm$0.01}&\boldmath\textbf{0.97$\pm$0.01}&\boldmath\textbf{0.98$\pm$0.01}&\boldmath\textbf{0.96$\pm$0.02}&\boldmath\textbf{0.97$\pm$0.02}\\
\multirow{2}{*}{{$\rm GraphSAGE$}}&{$\rm Imbalance$}&{$\rm 0.93$$\pm$0.02}&{$\rm 0.74$$\pm$0.15}&{$\rm 0.74$$\pm$0.13}&{$\rm 0.75$$\pm$0.20}&{$\rm 0.92$$\pm$0.03}&{$\rm 0.96$$\pm$0.01}&{$\rm 0.95$$\pm$0.02}&{$\rm 0.96$$\pm$0.02}\\
&\boldmath\textbf{{$\rm Banlance$}}&\boldmath\textbf{0.99$\pm$0.01}&\boldmath\textbf{0.98$\pm$0.01}&\boldmath\textbf{0.98$\pm$0.01}&\boldmath\textbf{0.98$\pm$0.01}&\boldmath\textbf{0.99$\pm$0.01}&\boldmath\textbf{0.98$\pm$0.01}&\boldmath\textbf{0.98$\pm$0.01}&\boldmath\textbf{0.99$\pm$0.01}\\
\bottomrule
\end{tabular*}
\label{tab:four}
\end{table*}

Tab .\ref{tab:fea} illustrates the training results of generating graph data using different feature generation methods. It is evident that MFG outperforms the conventional linear method. This superiority is particularly pronounced in the UNSW-NB15 dataset, where accuracy, recall, precision, and F1 score have each witnessed improvements of 0.79, 0.86, 0.86, and 0.93, respectively. Fig. \ref{fig:mutifea} presents feature visualization graphs for different methods, indicating that our approach can generate more complex patterns. On the contrary, the linear-based method produces data patterns with limited variations, showcasing more pronounced issues related to pattern collapse.

\begin{table*}[htbp]
\footnotesize
\renewcommand{\tablename}{Table}
\centering
\caption{Classification performance with different adjacency matrix sparsity.}
\begin{tabular*}{\linewidth}{@{\extracolsep{\fill}}cccccc@{}}
      \toprule
{$\rm Data$}&{$\rm Sparsity$}&{$\rm Accuracy$}&{$\rm Recall$}&{$\rm Precision$}&{$\rm F1-score$}\\
\midrule
\multirow{4}{*}{{$\rm UNSW-NB15$}}{}&{$\rm 0.700$}&{$\rm 0.40$$\pm$0.12}&{$\rm 0.36$$\pm$0.13}&{$\rm 0.34$$\pm$0.11}&{$\rm 
 0.34$$\pm$0.12}\\
&{$\rm 0.400$}&{$\rm 0.70$$\pm$0.10}&{$\rm 0.70$$\pm$0.08}&{$\rm 0.72$$\pm$0.08}&{$\rm 0.71$$\pm$0.08}\\
&\boldmath\textbf{ 0.142}&\boldmath\textbf{0.98$\pm$0.01}&\boldmath\textbf{0.98$\pm$0.01}&\boldmath\textbf{0.98$\pm$0.01}&\boldmath\textbf{0.98$\pm$0.01}\\
\\
\multirow{4}{*}{{$\rm CICIDS-2017$}}&{$\rm 0.700$}&{$\rm 0.49$$\pm$0.14}&{$\rm 0.45$$\pm$0.07}&{$\rm 0.48$$\pm$0.07}&{$\rm 0.44$$\pm$0.12}\\
&{{$\rm 0.400$}}&{$\rm 0.72$$\pm$0.05}&{$\rm 0.73$$\pm$0.04}&{$\rm 0.74$$\pm$0.05}&{$\rm 0.73$$\pm$0.03}\\
&\boldmath\textbf{ 0.138}&\boldmath\textbf{0.97$\pm$0.02}&\boldmath\textbf{0.98$\pm$0.01}&\boldmath\textbf{0.98$\pm$0.01}&\boldmath\textbf{0.98$\pm$0.01}\\
\bottomrule
\end{tabular*}
\label{tab:fir}
\end{table*}

\begin{figure*}[htbp]
\centering  
\subfloat[Training results in the UNSW-NB15.]{\label{fig:subfig1}\includegraphics[width=0.8\linewidth]{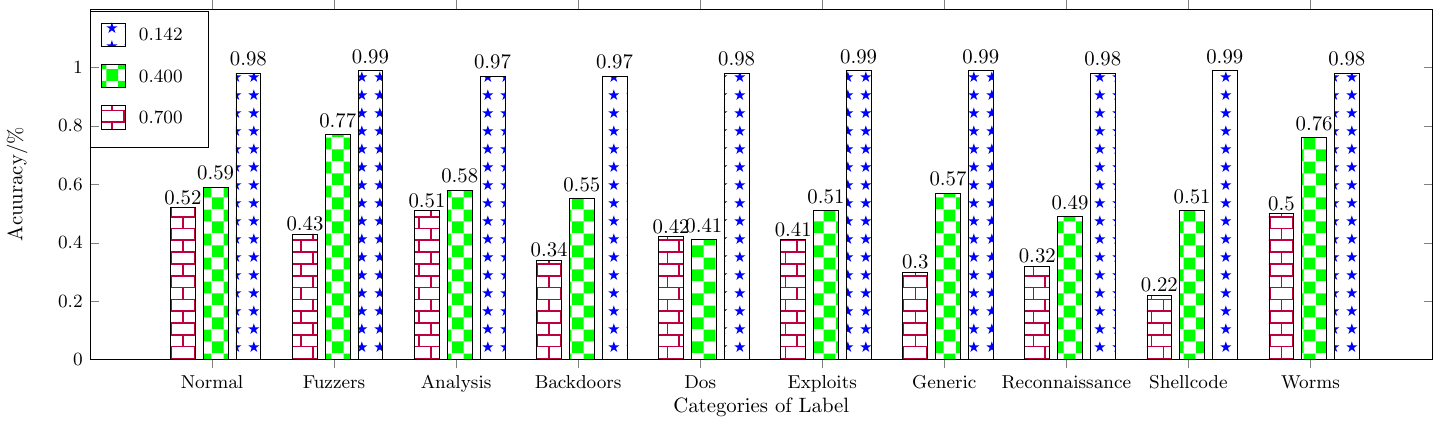}}\\
\subfloat[Training results in the CICIDS-2017.]{\label{fig:subfig2}\includegraphics[width=0.8\linewidth]{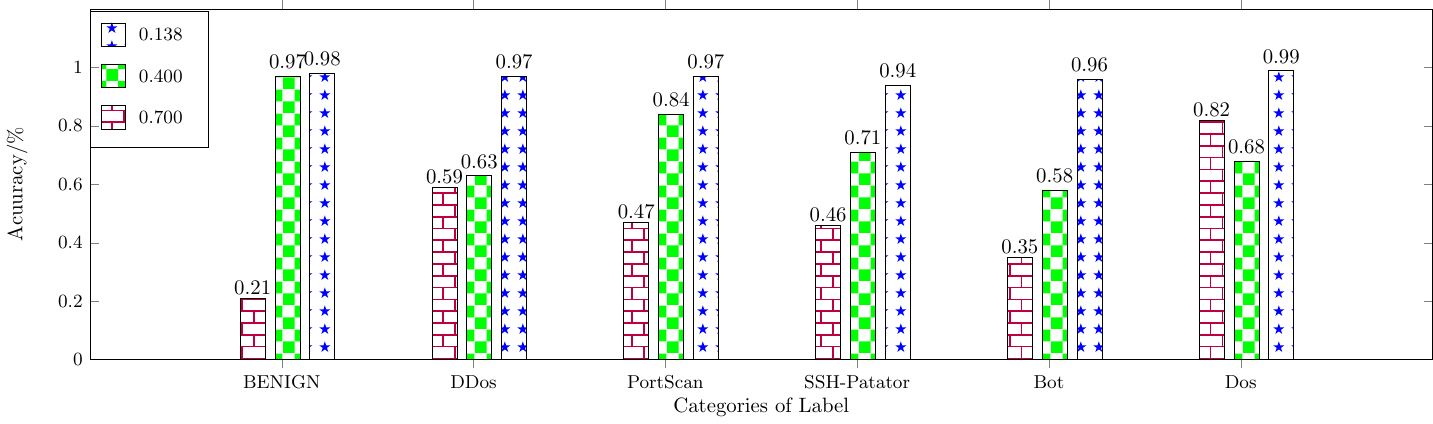}} \\
	\caption{Comparison performance with different
adjacency matrix sparsity} 
 \label{fig:tal}
\end{figure*}

\subsubsection{Different Classifiers Analysis}

To validate the generalization of the method proposed in this paper, we selected two graph neural networks as classifiers, namely  GCN (\cite{kipf2016semi}) and GraphSAGE (\cite{hamilton2017inductive}), for the anomaly detection task. To assess the effectiveness of the generated graph snapshots, we progressively trained an anomaly detection model on each generated snapshot. We operated under the assumption that the synthesized graph should mirror the fundamental characteristics of the real graph. we expect the anomaly detection classifier trained on synthetic data to effectively detect anomalies in real data. For evaluation, we utilized both real and synthetic data as training and testing datasets. The detailed results of the experiment are presented in Tab. \ref{tab:four}.

In most cases, the performance of the GCN model surpasses that of GraphSAGE, indicating that the GCN model is better suited for node anomaly detection tasks. It is also evident that a balanced dataset significantly enhances classification for each category. In the UNSW-NB15 dataset, the model's accuracy increased from 0.725 to 0.954 and from 0.717 to 0.916, respectively. Similarly, in the CICIDS-2017 dataset, the accuracy of the GCN and GraphSAGE models increased from 0.712 to 0.827 and from 0.712 to 0.802, respectively.

\subsubsection{Adjacency Matrix Sparsity Analysis}

We define matrix sparsity as the ratio of non-zero elements to the total number of matrix elements. According to our measurements, the average matrix sparsity in the two datasets is 0.142 and 0.138, respectively. Therefore, during the downsampling of the adjacency matrix noise $A_o$, we establish three sampling criteria to acquire three adjacency matrices with varying sparsity levels. Synthetic data is utilized as the training set, while real data serves as the test set. Tab. \ref{tab:fir} shows the classification performance of synthetic data with different sparsity adjacency matrix. Fig. \ref{fig:tal} reveals that synthetic data effectively mimics real data when the synthetic adjacency matrix closely aligns with the sparsity of the real adjacency matrix. In both datasets, a greater difference in sparsity between the generated and real adjacency matrices results in poorer performance of the model trained on the synthetic data. Experimental findings delineate the correlation between sparsity and classification performance, confirming the effectiveness of the adaptive sparsity adjacency matrix generation mechanism.

\section{Conclusions}
\label{sec:conclusion}
In this paper, we propose a graph generation model CGGM to generate graph snapshots with multi-category labels by introducing conditional constraints, and the proposed model is applied to the IoT anomaly detection. Then, extensive experiments compare the quality of data generated by CGGM with other data generation models such as CTGAN and TableGAN. The results show that the synthetic data generated by CGGM performs best with the real data in several similarity matrices. The results of model training based on different synthetic data show that the synthetic data generated by CGGM can distinguish between different node classes to the most extent, which can significantly improve the classification performance and is more suitable for traffic-based anomaly detection tasks.

CGGM provides a promising approach for graph generation by combining attribute and structure learning. In future work, we'd like to extend the current offline-trained CGGM model to a real-time learning model. Additionally, we will explore deploying CGGM to real industrial IoT environments. The training and testing will not be limited to the public datasets, but also interacting with real-time traffic data, to evaluate CGGM's performance and adaptability in handling complex industrial data.

\section*{Declaration of Competing Interest}
The authors declare that they have no known competing financial interests or personal relationships that could have appeared to influence the work reported in this paper.

\section*{Acknowledgment}
This work was supported by the Liaoning Natural Science Funds (Grant No. 2024-BS-015), and in part by China Postdoctoral Science Foundation (No.2024M750295).

\bibliographystyle{elsarticle-num}
\balance
\bibliography{egbib}
~~~\\
~~~\\







\end{document}